\documentclass[11pt]{article}
\usepackage{amsmath,  amssymb}
\usepackage{mathrsfs}
\usepackage{amsthm}
\usepackage{multirow}
\usepackage{multicol}
\usepackage{array}
 
\usepackage{graphicx, color, epsfig}
\usepackage{wrapfig}
\usepackage{colortbl}

\newcommand{\R}{{\mathbb R}} 
\newcommand{\fR}{{\mathscr R}}

\newcommand{\N}{{\mathbb N}}

\newcommand{\grad}{{\mathop {\rm grad}}}

\newcommand{\vvte}{{\mathbb \theta}} 
\newcommand{\te}{{\theta}}
\newcommand{\tea}{{{\theta_1}}}
\newcommand{\tee}{{{\theta_2}}}
\newcommand{\that}{{{\hat{\theta}}}}

\newcommand{\deq}
   {\mathrel{\stackrel{\mathop {\rm def}}{=}}}

\newcommand{\var}{{\mathop {\rm Var}}}
\DeclareMathOperator*{\poly}{poly}

\newtheorem{theorem}{Theorem}[section]
\newtheorem{lemma}[theorem]{Lemma}

\newtheorem{cor}[theorem]{Corollary}
\newtheorem{proposition}[theorem]{Proposition}

\newtheorem{defn}[theorem]{Definition}

\def\beq{\begin{equation}}
\def\eeq{\end{equation}}

\def\bp{\begin{proof}}
\def\ep{\end{proof}}

\def\bl{\begin{lemma}}
\def\el{\end{lemma}}

\def\bt{\begin{theorem}}
\def\et{\end{theorem}}

\def\bprop{\begin{proposition}}
\def\eprop{\end{proposition}}

\def\bcor{\begin{cor}}
\def\ecor{\end{cor}}

\def\beq{\begin{equation}}
\def\eeq{\end{equation}}

\def\bl{\begin{lemma}}
\def\el{\end{lemma}}
\def\bt{\begin{theorem}}
\def\et{\end{theorem}}
\newcommand\beginproof{\noindent {\sc Proof:}}
\def\bp{\beginproof}
\def\ep{\hfill $\Box$}

\begin{document}

\title{Polynomial Learning of Distribution Families}

\author{Mikhail Belkin\\
Ohio State University \\
Columbus, Ohio\\
\texttt{\small mbelkin@cse.ohio-state.edu}
\and Kaushik Sinha\\
Ohio State University \\
Columbus, Ohio \\
\texttt{\small sinhak@cse.ohio-state.edu}
}


\maketitle

\begin{abstract}

The question of polynomial learnability of probability distributions, particularly Gaussian mixture 
distributions, has recently received significant attention in  theoretical computer science and machine learning. 
However, despite major progress, the general question of polynomial learnability  of Gaussian mixture 
distributions
still remained open. The current work  resolves the question of polynomial learnability for Gaussian mixtures in 
high dimension with  an arbitrary fixed number of components.

The result on learning Gaussian mixtures relies on an analysis  of distributions 
belonging to what we call {\it polynomial families} in low dimension. 
These families are characterized by their moments being 
polynomial in parameters and  include almost all common probability 
distributions as well as their mixtures and products. 
Using tools from real algebraic geometry, we show that parameters of any distribution belonging to 
such a family can be learned in polynomial time and using a polynomial number of sample points. 
The result on learning polynomial families is quite general and is of independent interest.

To estimate parameters of a Gaussian mixture distribution in high dimensions, we provide  
a  deterministic algorithm for dimensionality reduction. This allows us to reduce learning a 
high-dimensional mixture to a polynomial number of parameter estimations in low dimension.
Combining this reduction with the results on polynomial families yields our result on learning 
arbitrary Gaussian mixtures in high dimensions.

%



\end{abstract}

\section{Introduction}

Estimating parameters of a model from sampled data is one of the oldest and most general problems of statistical 
inference. Given a number of samples, one needs to choose a distribution that 
best fits the observed data. While traditionally theoretical analysis in the statistical literature has 
concentrated on rates (e.g., minimax rates), in recent years other computational aspects of this problem, 
especially as dependence on dimension of the space, have 
attracted attention.  In particular,  a recent line of work in the theoretical computer 
science and learning communities has been 
concerned with learning the distribution in time and using the number of samples,  polynomial in parameters 
and the dimension of the  space. This effort has been particularly directed at the family of Gaussian Mixture 
models due to their simple formulation and widespread use in applications spanning areas such as computer vision, 
speech 
recognition, and many others (see, e.g.,\cite{lindsay95, mclachlan88, titter85}).
This line of research   started with the  work of Dasgupta~\cite{dasgupta99}, who was the
first to show that learning the parameters of a Gaussian mixture distribution in time polynomial 
in the  dimension of the space $n$ was possible at all. 
This work has been 
refined and extended in a number of consequent papers. The results in~\cite{dasgupta99} required separation 
between mixture components on the order of $\sqrt{n}$. That was later improved 
to  
of $\Omega(n^{\frac{1}{4}})$
in~\cite{dasgupta00} for mixtures of spherical Gaussians and in~\cite{arora01} 
for general Gaussians. The separation 
requirement was
further reduced and made independent of $n$ to the order of $\Omega(k^{\frac{1}{4}})$ in \cite{vempala02} for 
a mixture of $k$ spherical
Gaussians and to the order of $\Omega(\frac{k^{\frac{3}{2}}}{\epsilon^{2}})$ in~\cite{kannan05} for logconcave 
distributions. In ~\cite{achlioptas05} the separation requirement was further reduced to 
$\Omega(k+\sqrt{k\log n})$.
An extension of PCA called isotropic PCA  was introduced in \cite{brubaker08}  to 
learn mixtures of Gaussians when any pair of
Gaussian components is separated by a hyperplane having very small overlap along the hyperplane direction 
(so-called "pancake layering problem").
A number of recent papers~\cite{chaudhuri09, chaudhuri08a, chaudhuri08b, dasgupta05, feldman08} addressed related problems, such as learning mixture
of product distributions and heavy tailed distributions.

\begin{table}[t]
\centering
\begin{tabular}{|l|c|l|}\hline
~~~~~~~~~~~~~~~~~Author & Min. Separation & ~~~~~~~~~~~~~~~~~~~~~~~Description\\ \hline
Dasgupta \cite{dasgupta99} & $\sqrt{n}$ & Gaussian mixtures, mild assumptions \\ \hline
Dasgupta-Schulman \cite{dasgupta00} & $n^{\frac{1}{4}}$ & Spherical Gaussian mixtures\\ \hline
Arora-Kannan \cite{arora01} & $n^{\frac{1}{4}}$ & Gaussian mixtures\\ \hline
Vempala-Wang \cite{vempala02} & $k^\frac{1}{4}$ & Spherical Gaussian mixtures\\ \hline
Kannan-Salmasian-Vempala \cite{kannan05} & ${k^\frac{3}{2}}$ & Gaussian mixtures, log-concave distributions\\ \hline
Achlioptas-McSherry \cite{achlioptas05}& $k+\sqrt{k\log n}$ & Gaussian mixtures \\ \hline
Feldman-Servedio-O'Donnell \cite{feldman06}& $>0$ & Axis aligned Gaussians, no param. estimation \\ \hline
Belkin-Sinha \cite{belkin09} & $>0$  & Identical spherical Gaussian mixtures\\ \hline
Kalai-Moitra-Valiant \cite{kalai10} & $\ge0$  & Gaussians mixtures with two components \\ \hline
{\bf This paper} & $\ge0$ & Gaussian mixtures\\ \hline
\end{tabular}
\label{tab:separation_result}
\caption{Partial summary of results on Gaussian mixture model learning. Note that~\cite{feldman06}
addresses a somewhat different problem.
 The last two methods allow the separation between the means to be zero, assuming different covariance matrices.}
\end{table}

However all of these papers assumed a minimum separation between the components, 
which is an increasing function of   the dimension  $n$ and/or the number of components $k$. 
The general question of  learning  parameters of a distribution without any separation conditions, remained open. 
The first result in that 
direction was obtained in Feldman, et al.,~\cite{feldman06}, which showed that the density (but not the parameters) 
of mixtures of axis aligned Gaussians can be learned in polynomial time using the method of moments.

Very recently two papers~\cite{belkin09,kalai10} independently  addressed  two special cases of Gaussian mixture 
learning without separation assumption. 
In Kalai, et al.,~\cite{kalai10} the authors 
showed that a mixture of two Gaussians  with arbitrary covariance matrices can be learned in polynomial time. The 
technique relies on a randomized algorithm to reduce the problem to one dimension. The key argument of the paper 
is  based  on  deconvolving the one-dimensional mixture to increase the separation between the components and 
carefully analyzing the moments of the deconvolved mixture in order to apply the method of moments.
In~\cite{belkin09} it is shown that a mixture of $k$ identical spherical  Gaussians can be learned in time 
polynomial in  dimension.  The key result is based on analyzing the Fourier transform of the 
distribution in one dimension  to give  a lower bound on the norm. 
However, it is not clear whether the techniques of either~\cite{kalai10} or~\cite{belkin09} 
could be applied to the general case 
with an arbitrary number of components and covariance matrices.

In this paper we resolve the polynomial learnability problem by proving that there exists 
a polynomial  algorithm to estimate parameters of a general high-dimensional mixture with arbitrary fixed 
number of Gaussians components without any  additional assumptions. Table~\ref{tab:separation_result} briefly 
summarizes the progress in the area and our result.

Our main result for Gaussian mixtures relies on a quite general result of independent interest on learning 
what we call \emph{polynomial families}. These families are characterized  by their moments being
polynomial in the parameters of a distribution. It turns out that almost all common 
distribution families, e.g., Gaussian, exponential, uniform, Laplace, binomial, Poisson and a number of others.  (see  Table~\ref{tab:poly_table}  in Appendix~\ref{app:distribution_family} for a longer list and a description of their moments),  as well as their 
mixtures and (tensor) products have this property.  Our technique uses methods of real algebraic geometry and 
combines them with the  classical method of moments (originally introduced by Pearson in~\cite{pearson1894} to 
analyze Gaussian mixtures).  

We note that there  have been  applications of algebraic geometry in the field of statistics, particularly in
conditional independence testing and likelihood estimation for discrete distributions and  exponential families 
(see, e.g.,~\cite{drton09}). We 
note that a mixture of more than one Gaussian distributions is a family of continuous distributions, which is not 
an exponential family.

%

Below we give a brief summary of the main results and the structure of the paper.

\noindent{\bf Brief outline of the paper.}\\
{\bf Section~\ref{sec:poly_families}.} We start Section~\ref{sec:poly_families} by introducing the problem of
parameter learning and defining the notion of a {\it polynomial family}. 
We proceed to prove the main result showing that parameters of a distribution
from a polynomial family can be learned with confidence  $1-\delta$ up to precision $\epsilon$ using the number 
of samples $poly(\frac{1}{\delta}, \max(\frac{1}{\epsilon}, \frac{1}{\fR}))$, where $\fR$ is the {\it radius of 
identifiability}, a measure of intrinsic hardness of unique parameter identification for a distribution\footnote{For 
example, it is impossible to identify mixing coefficients of a mixture of two Gaussians with identical means and 
variances, thus in that case $\fR=0$. See Section~\ref{sec:poly_reducible} for the detailed analysis of Gaussian mixtures.}. In fact, the result is more general, even if the radius of identifiability is zero, parameters can 
still be learned up to a certain equivalence relation defined in the paper.
 
The proof consists of the two main steps.   The first step uses the  Hilbert basis theorem 
for an appropriately defined ideal in the ring of  polynomials to show that a fixed set of (possibly 
high-dimensional) 
moments uniquely identifies the distribution.

In the second step, we pose parameter estimation problem as a system of quantified algebraic equations and 
inequalities using the finite set of moments obtained in the first step.  We use quantifier elimination  for 
semi-algebraic sets (Tarski-Seidenberg theorem) to prove  that there exists a polynomial algorithm for parameter 
learning. 

\noindent{\bf Section~\ref{sec:poly_reducible}.} In Section~\ref{sec:poly_reducible} 
we prove our main results on learning Gaussian mixture distributions in high dimensions. 
The main difficulty is that the general results of Section~\ref{sec:poly_families} cannot be applied directly since 
the number of parameters increases with the dimension of the space. To overcome this issue, we prove that the 
Gaussian family has the property that we call {\it polynomial reducibility}. That is the parameters of a distribution 
in $n$ dimensions can be recovered from a $poly(n)$ number of low-dimensional projections. Specifically, we show
that a mixture of Gaussians with $k$ components can be recovered using a polynomial number of projections to 
$(2k^2+2)$-dimensional space. This leads us to Theorem~\ref{thm:gaussian_learning}, our main result for parameter learning on Gaussian mixtures. We show 
that parameters of a Gaussian mixture can be learned with precision $\epsilon$ and confidence $1-\delta$, 
using the number of samples  polynomial in dimension $n$, $\frac{1}{\delta}$ and 
$\max(\frac{1}{\epsilon},\frac{1}{\fR})$. Moreover, we also provide explicit formula for the radius of identifiability of Gaussian mixtures. If we are given an a priori bounds on the minimum mixing weight 
and the minimum separation between the mean/covariance pairs, that leads to an upper bound on $\frac{1}{\fR}$. 
For example, our results holds even in the extreme case where all components have the same mean, as long as the covariance matrices are different. In Theorem~\ref{thm:identifiability_detection} we also show that in the absence of such a lower bound, 
$\fR$ can be estimated directly from the data. 

We discuss other polynomially reducible families, where a 
similar approach would  yield results on polynomial learnability.

%
%

In {\bf Section~\ref{sec:conclusion}} we conclude and discuss some limitations of our results, directions 
of future work and conjectures.



\section{Learning Polynomial Families} \label{sec:poly_families}

In this section we prove some general learnability results for  a large class of probability distributions that we 
call {\it polynomial families}, which are characterized  by the moments being polynomial functions of parameters.
This class  turns out to contain nearly all commonly used probability distributions, as well as their mixtures 
and (tensor) products.  
See Appendix~A (Table~\ref{tab:poly_table}) for a partial list together with the description of their moments either explicitly or through a recurrence relation, as well as some examples of families, which are not polynomial (Table~\ref{tab:non_polynomial}).

\begin{wrapfigure}{l}{0.47\textwidth}
\vspace{-0.3in}
\centerline{\includegraphics[scale=0.32]{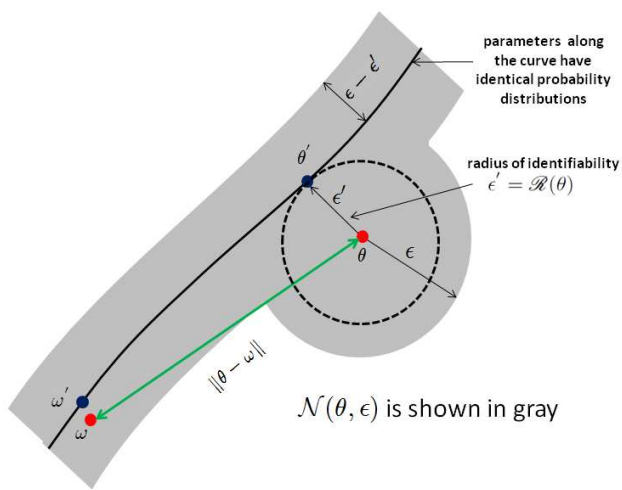}}
\vspace{-0.5in}
\caption{\small If $\theta$ and $\omega$ are close to two values of
parameters $\theta'$ and $\omega^{'}$ with identical
probability distribution, then it is be hard to distinguish between
them from sampled data, even when $\|\theta-\omega\|$ is large.}
\label{fig:fiber}
\vspace{-0.3in}
\end{wrapfigure}

The main result in this section is Theorem~\ref{thm:existence_of_poly_algo}, which shows that there exists an 
algorithm to  learn the parameters of a polynomial distribution using a polynomial number of samples. 

We start with the outline of the standard
parameter learning problem.
Let $p_\vvte$, $\vvte=(\theta^1,\ldots,\theta^m), \theta \in \Theta \subset \R^m$  be a $m$-parametric family of 
probability distributions in $\R^l$.  The problem of parameter learning is the following: given precision 
$\epsilon$ and confidence $\delta$, and some number $n(\epsilon, \delta)$ of points sampled from $p_\te$,
we need to provide an estimate $\hat{\theta}$, such that  $\|\hat{\theta} - \theta\| < \epsilon$ with probability 
at least $1-\delta$.

However, for many families identifying the values of 
parameters  uniquely is impossible, due to the fact that several different values of parameters 
may correspond to the same  probability distribution. 
Moreover, if two values of parameters, say, $\te$ and $\omega$ are close to two values
of parameters, $\theta'$ and $\omega'$ respectively, which have identical probability distributions, then it may be 
hard  to distinguish between  them. This situation is illustrated in Fig.~\ref{fig:fiber}.  These observations
suggests that a more general formulation of learning distribution parameters needs to take these 
 into account.  A mathematical formalization of the more general of learnability will be given in
Eq.\ref{def_dist_distance}, which defines a notion of a neighborhood taking parameters with identical probability distribution into account. An $\epsilon$-''neighborhood'' of $\te$, ${\cal N}(\te,\epsilon)$, is shown in gray in Fig.~\ref{fig:fiber}.
We will also introduce the notion of the {\it 
radius of identifiability} $\fR(\theta)$ (definition~\ref{def:rad_identifiability}) to give a quantification of how hard it may be to identify the 
parameters. 
For example, parameters $\theta$ for which $\fR(\theta) = 0$ cannot be identified given any amount of data. 
In Fig.~\ref{fig:fiber}, the radius of identifiability $\fR(\te)$ is equal to $\epsilon'$.


For mixtures of Gaussians any permutation of the mixture components has the 
same distribution, while a component with zero mixing weight may have arbitrary mean/covariance. If two components have the same mean/covariance pair, 
then the mixing coefficients are not defined uniquely. 
However, assuming that the mean/variance pairs for any two components are different and that the
mixing coefficients are  non-zero, the parameters are defined uniquely up to a permutation of components 
(see Section~\ref{sec:poly_families}).

Our main Theorem~\ref{thm:existence_of_poly_algo} applies even when parameters of a probability distributions are 
not defined uniquly, including 
the standard definition of parameter learning  as a special case (see Corollary~\ref{cor:general_case_0} and 
Corollary~\ref{cor:general_case_1}).

In Subsection~\ref{subsec:poly_families} we prove the basic properties of polynomial families, including the key 
result, Theorem~\ref{th_identifiability}, which shows that a finite set of moments  uniquely determines the 
distribution.

In Subsection~\ref{subsec:learning_poly_families} we define the extended notion of a neighborhood  ${\cal N}(\te, 
\epsilon)$ and discuss its basic properties. We proceed to obtain the main technical result, a lower bound in 
Theorem~\ref{th_lower_bound}. This, together with the upper bound in Proposition~\ref{prop_upper_bound} allows us to set up a 
grid search to prove the main Theorem~\ref{thm:existence_of_poly_algo}. We also define the radius of identifiability, and derive Corollary~\ref{cor:general_case_0} and Corollary~\ref{cor:general_case_1}.


\subsection{Polynomial Families and Finite Sets of Moments}\label{subsec:poly_families}

We  start by assuming that the parameter set $\Theta$ is a compact semi-algebraic subset of $\R^m$.
Recall that a semi-algebraic set in $\R^m$ is a finite union of sets defined 
by a system of algebraic equations and inequalities. 
A sphere, a polytope, the sets of symmetric and orthogonal matrices  are all examples of 
semi-algebraic sets. For example, a typical family of Gaussian mixture distributions with bounded means and bounded (in norm) covariance matrices would satisfy this condition.

The family of semi-algebraic sets is closed under finite union,  intersection and taking complements.
Importantly, the Tarski-Seidenberg theorem states that a linear projection of a semi-algebraic set is also 
semi-algebraic. This is  equivalent  to the \emph{elimination of quantifiers} for semi-algebrac sets, which we will need shortly. See~\cite{basu03} for a review of results on real algebraic geometry.

\begin{defn}  [Polynomial family]
We call the family $p_\theta$ a polynomial family, if each (raw $l$-dimensional) moment $M_{i_1,\ldots,i_l}(\theta) = \int x_1^{i_1}\ldots x_l^{i_l} \,d p_\te$  of  the distribution exists and can be represented as a polynomial of the parameters 
$(\theta^1,\ldots,\theta^m)$.  We also require that each $p_\theta$ should be defined uniquely by its moments\footnote{This is true under some mild conditions, e.g., if the moment generating function converges in a neighborhood of zero~\cite{feller71}.}.
\end{defn} 
We will order the moments $M_{i_1,\ldots,i_l}$ lexicographically and denote them by $M_1(\te),\ldots,M_n(\te),\ldots$ 
In the one-dimensional case this corresponds to the standard ordering of the moments.

As it turns out, most of the  
common families of probability distributions are, in fact,  polynomial (see Appendix~A).
Moreover,  a mixture, a product or a linear transformation of polynomial families is also a polynomial family, as stated in the following
\bl Let $p_\theta$, $\theta \in \Theta$  and $q_\omega$, $\omega \in \Omega$ be polynomial families. Then the following families are also polynomial:\\
(a) the family 
$ w_1 p_\theta + w_2 q_\omega$, $w_1,w_2 \in \R$, $w_1 + w_2 = 1$.\\
(b) the family $p_{\te,\omega}(x,y) = p_\theta(x)\times p_\omega(y)$, $(\te,\omega) \in \Theta \times \Omega$. \\
(c) the family $p_{A\theta}$, where $A\in\R^{m\times m}$ is a fixed matrix and $A\theta\in \Theta \subset \R^m$.
\el
The proof follows directly from the linearity of the integral, the 
Fubini's theorem and the fact that polynomial functions stay polynomial under a linear change. 


Note that a multivariate Gaussian distribution is a product  of univariate Gaussians along its principal directions of the covariance matrix. Since the standard coordinates can be transformed to principal coordinates by a linear transformation, a multivariate Gaussian is a polynomial family. 
Hence a general mixture of $k$ multivariate normal distributions in $\R^l$ is also a polynomial family with $lk + \frac{1}{2}l(l+1)k + k-1$ parameters.

Let us now recall that a family $p_\theta$, is called \emph{identifiable} if $p_{\tea} \ne p_{\tee}$  for any $\tea \ne \tee$.
We will  now prove the following
\bt  \label{th_identifiability}
Let $p_\theta$ be a polynomial family of distributions. Then  there exists a positive integer $N$, such that $p_\tee = p_\tea$ 
if and only if $M_i(\tea) = M_i(\tee)$ for all $i=1,\ldots,N$. In the case when the family $p_\theta$ is identifiable, the first $N$ 
moments are 
sufficient to uniquely identify the parameter $\theta$.
\et
\bp \\
Since Let $p_\te$ is a polynomial family, each $M_i(\te)$ is a polynomial 
of $\te$. Let $\tea = (\te_1^1,\ldots, \te_1^m)$ and $\tee = (\te_2^1,\ldots, \te_2^m)$. Let 
$$P_i(\te_1^1,\ldots, \te_1^m, \te_2^1,\ldots, \te_2^m) = M_i(\tea) - M_i(\tee)$$ 
be a polynomial of $2m$ variables.
Now let ${\cal I}_j$ be the ideal in the ring of polynomials of $2m$ variables generated by the polynomials $P_1,\ldots,P_j$. 
Thus we have an increasing sequence of ideals ${\cal I}_1 \subset {\cal I}_2 \subset {\cal I}_3 \ldots$  Let ${\cal I} = \cup_{j=1}^\infty~ {\cal I}_j$.
By the Hilbert basis theorem, the ideal $\cal I$ is finitely generated, which implies that for some $N$ large enough, ${\cal  I}_N$  contains all of the 
generators.  Therefore for any $M \ge N$ we can write 
$$
P_M (\tea, \tee) = \sum\limits_{i=1} ^N a_i(\tea, \tee) P_i(\tea, \tee)
$$
for some polynomials $a_i$.
Thus if $P_i(\tea, \tee)=0$ for $i=1,\ldots,N$ then $P_i (\tea, \tee)=0$ for any $i$.
Recalling the definition of $P_M$, we conclude that all moments of $p_\tea$ and $p_\tee$ coincide if and only if the first 
$N$ moments of these distributions are the same. Since the sequence of moments defines the distribution uniquely, the statement of the theorem follows.\ep

\subsection{Learning Polynomial Families}\label{subsec:learning_poly_families}
We  will now  introduce a notion of an $\epsilon$-''neighbourhood'' of a point, which takes into account that 
different parameters may have identical probability distribution. We proceed to prove the main Theorem~\ref{thm:existence_of_poly_algo} and a few corollaries, showing that the standard parameter learning problem becomes a special case of the result.

Let ${\cal E}(\theta) =  \{\omega | p_\omega = p_\theta\}$  be the set of parameters $\omega$  which  have  distributions same as  $p_\theta$. 
We note that the distributions corresponding to different values of parameters in the set ${\cal E}(\theta)$ are identical and hence cannot be 
distinguished  from  each other given any amount of sampled data.
We now define 
\begin{equation}\label{def_dist_distance}
{\cal N}(\theta, \epsilon) = \{\omega \in \Theta |\, \exists_{\omega',\te' \in \Theta, 0<\epsilon'<\epsilon}~ \|\omega - \omega'\| <\epsilon', \omega' \in {\cal E}(\theta'), \|\te' - \theta\| <\epsilon-\epsilon'\}
\end{equation}
In other words, $\omega$ belongs to ${\cal N}(\theta, \epsilon)$ if it is within $\epsilon' < \epsilon$ distance of a parameter value which 
has the same probability distribution as a parameter value within $\epsilon - \epsilon'$ of $\theta$.  This definition is illustrated graphically in 
Fig.~\ref{fig:fiber}.
We observe the following properties of ${\cal N}(\theta, \epsilon)$:
\begin{enumerate}
\item (Symmetry)  If $\tea \in {\cal N}(\tee, \epsilon)$ then $\tee \in {\cal N}(\tea, \epsilon)$.
\item ($\epsilon$-ball) An $\epsilon$-ball $B(\theta, \epsilon)$ around $\theta$ is contained in   ${\cal N}(\theta, \epsilon)$. If   
$B(\theta, \epsilon)$ is an identifiable family, then  $B(\theta, \epsilon) = {\cal N}(\theta, \epsilon) $.
\item (Equivalence) If $p_\tea = p_\tee$, then $\tea \in {\cal N}(\tee, \epsilon)$ for any $\epsilon >0$.
\end{enumerate}

Thus ${\cal N}(\theta, \epsilon)$ can be viewed as an ``$\epsilon$-ball" around 
$\theta$ taking probability distribution into account. 
For example, values of parameters with identical probability distributions cannot be distinguished by this metric, which is consistent with statistical identifiability.

\bl \label{lemma_open}
${\cal N}(\theta, \epsilon)$ is an open semi-algebraic set.
\el
\bp
${\cal N}(\theta, \epsilon)$ is open since, a sufficiently small open ball around any point $\omega \in {\cal N}(\theta, \epsilon)$ is also contained in ${\cal N}(\theta, \epsilon)$. 
To see that it is algebraic we recall that by Theorem~\ref{th_identifiability} there exists an $N$, such that $\tea \in {\cal E}(\tee)$ if and only if 
\beq
Q(\tea,\tee) \deq \sum_{i=0}^N (M_i(\tea) - M_i(\tee))^2 = 0
\eeq
which is an algebraic condition. Hence, by applying the Tarski-Seidenberg theorem to eliminate the existential quantifiers in Eq.~\ref{def_dist_distance}, we see that ${\cal N}(\theta, \epsilon)$ is semi-algebraic.\ep

\bt [Lower bound] \label{th_lower_bound} Let $p_\te$ be a  polynomial family. 
There exists  $N \in \N$ and $t>0$, such that for any sufficiently small $\epsilon >0$ and any $\tea, \tee \in \Theta$, if $|M_i(\tea) - M_i(\tee)| > \epsilon$   for at least one $i \le N$, then $\tea \notin {\cal 
N}(\tee, O(\epsilon^t))$.
\et
\bp \\
Choose $N$ as in Theorem~\ref{th_identifiability}.  We start by observing we can replace the condition 

$|M_i(\tea) - M_i(\tee)| > \epsilon$ by 
$$
Q(\tea,\tee) \deq \sum_{i=1}^N |M_i(\tea) - M_i(\tee)|^2 >N{\epsilon^2}
$$ in the statement of the theorem. Since the existence of $t$ is not affected by the substitution of $N{\epsilon^2}$, instead of 
$\epsilon$, to simplify the matters we will  assume  that $Q(\tea,\tee) > \epsilon$. 

From Theorem~\ref{th_identifiability} we recall that if for some $i \le N$ $|Q(\tea,\tee)| \ne 0$ then $p_\tea \ne p_\tee$. 
Let $\delta$ be a positive real number.  
Consider the set $X = \{\tea, \tee | \tea \in {\cal  N}(\tee, \delta)\}$. From Lemma~\ref{lemma_open} and the fact that the relationship $\tea \in {\cal  N}(\tee, 
\delta)$ is symmetric, it follows that $X$ is an open subset of $\Theta \times \Theta$. Hence the set $\Theta \times \Theta - X = \{\tea, \tee \in \Theta, \tea \notin {\cal  N}(\tee, \delta))\}$ is compact
and since $Q(\tea, \tee) >0$ for any $(\tea,\tee) \in  \Theta \times \Theta - X$ we have
\begin{equation}\label{eq_inf}
\inf_{\tea, \tee \in \Theta, \tea \notin {\cal  N}(\tee, \delta))}  Q(\tea, \tee) >0
\end{equation}
By an argument following that in Lemma~\ref{lemma_open} we see that $X$ and hence its complement are semi-algebraic sets.

%
%

Consider now the set $S_\delta$, $\delta >0$  given by the following expression 
\beq
S_\delta = \{\epsilon >0  \, | \,   \forall_{{\tea, \tee \in \Theta}} ~~ (\tea \notin {\cal  N}(\tee,\delta)) \Rightarrow  Q(\theta_1, \theta_2) >\epsilon \}.
\eeq

Since these logical statements can be expressed as semi-algebraic conditions, by the Tarski-Seidenberg theorem $S_\delta$ is a 
semi-algebraic subset of $\R$. Let $\epsilon(\delta) = \inf S_\delta$.
From  Eq.\ref{eq_inf} we have that  $\epsilon(\delta) >0$ for any positive $\delta$. 
Since the number $\epsilon(\delta) >0$ is easily written using quantifiers and algebraic conditions, the
Tarski-Seidenberg theorem implies that it is a semi-algebraic set and hence satisfies 
some algebraic equation\footnote{Note that strict inequalities alone cannot define a set consisting of a single point.}  whose  coefficients  are polynomial in $\delta$.

We write this polynomial as 
$ q(x) = q_M(\delta) x^M + \ldots + q_0(\delta)$, such that $q(\epsilon(\delta))=0$.
We can assume that $q_0(\delta)$ is not identically zero (dividing by an appropriate power of $x$ if necessary).
From Lemma~\ref{lemma_rootbound} we see that if $q(\epsilon(\delta)) = 0$ then
$$
\epsilon(\delta) > \frac{|q_0(\delta)|}{\sum_{i=1}^M  |q_i (\delta)|}.
$$
The last quantity is a ratio of two polynomials in $\delta$ and can thus be lower bounded by $C(\delta^{t'})$, so that $\epsilon(\delta) > C \delta^{t'}$
for some $t'>0$, when $\delta$ is sufficiently small. 

Putting $t = \frac{1}{t'}$ and recalling the definition of $S_\delta$, we see that $Q(\theta_1, \theta_2) < \epsilon$, 
implies $\tea \in {\cal  N}(\tee, O(\epsilon^t)) $, which completes the proof of the theorem. \ep

\bl \label{lemma_rootbound} Let $\delta$ be a positive root of the polynomial $q(x) = a_M x^M+\ldots+a_0$, $a_0\ne 0$.
Then 
$
\delta > \min(\frac{\sum_{i=1}^M |a_i|}{|a_0|},1)
$.
\el
\bp
We have $\delta(\sum_{i=1}^M a_i \delta^{i-1}) = -a_0$. For $0<\delta<1$  we have $\sum_{i=1}^M a_i \delta^{i-1} < \sum_{i=1}^M |a_i|$, 
and the 
statement follows.\ep

\bprop [Upper bound] \label{prop_upper_bound} Let $p_\te$ be a  polynomial family. 
For any  $N \in \N$ there exists a $C >0$,  such that
$$
\sum_{i=1}^N |M_i(\tea) - M_i(\tee)|^2 < C \|\tea -\tee\|^2.
$$
If $\Theta$ is contained in a ball of diameter $B$, then    $C$ is bounded from above by a polynomial of $B$. 
\eprop
\bp 
To prove the claim it is sufficient to show that each summand $|M_i(\tea) - M_i(\tee)|^2$ is bounded from above by $C' \|\tea -\tee\|^2$, which is equivalent to proving that
$\frac{|M_i(\tea) - M_i(\tee)|}{\|\tea -\tee\|} < \sqrt{C'}$. We now observe  that by the mean value theorem
$$
\frac{|M_i(\tea) - M_i(\tee)|}{\|\tea -\tee\|} \le \sup_{\theta \in \Theta} \|\grad(M_i)(\te)\|
$$
where  $\grad$ is the gradient of the function $M_i$. Since $M_i$ is a polynomial, all elements of the vector $\grad(M_i)$ are polynomial in $\theta$. Therefore
$$\sup_{\theta \in \Theta} \|\grad(M_i)(\te)\| < C'' B^t$$ where $t$ is the maximum degree of these polynomials and $C''$  is an appropriate constant. This implies the statement of the Proposition.\ep

Now we have the following:
\bt\label{thm:existence_of_poly_algo}
There exists an algorithm, which,  given $\epsilon>0$ and $1>\delta>0$,   and   $P(\frac{1}{\epsilon},\frac{1}{\delta},B)$ 
samples from $p_\theta, \theta\in\Theta$, where $\Theta$ is the set of parameters within a ball of radius $B$ and $P$ is a polynomial depending only on the distribution family,
outputs $\hat{\theta}$, s.t. $\hat{\theta} \in {\cal  N}(\theta, \epsilon)$ with probability at least $1-\delta$. The algorithm also requires a polynomial number of operations.
\et
\bp From Theorem~\ref{th_lower_bound} it follows that there exists an $N \in \N$ and $t>0$, 
such that if $\forall_{i=1,\ldots,N} |M_i(\that) - M_i(\te)| < \epsilon^t$, than $\that \in {\cal N}(\te, \epsilon)$. 
Thus it is sufficient to estimate each moment within $O(\epsilon^t)$.  From Lemma~\ref{lem:moment_concentration} (moment estimation) this can be done with probability $1- \delta$ given a number of 
sample points $\poly(\frac{1}{\epsilon^t},\frac{1}{\delta} )= \poly(\frac{1}{\epsilon},\frac{1}{\delta})$ by computing the empirical moments of the sample. Once we 
have precise estimates of the first moments a simple grid search suffices to find the corresponding values of parameters. 
Indeed, suppose that $\Theta$ is contained in a ball of radius $B$ in $\R^m$.
Then the desired estimate can 
be obtained by conducting a grid search   
over a rectangular grid of size  $O(\frac{\epsilon^t}{N\sqrt{m}})$ and invoking Proposition~\ref{prop_upper_bound}.  We see that the number of operations is 
polynomial in $\epsilon$ and the main theorem is proved.\ep

To simplify further discussion  we will now define the {\it radius of identifiability}:
\begin{defn}\label{def:rad_identifiability} As before let $p_\theta$, $\theta \in \Theta$ be a family of probability distributions. For each $\theta$ we  define the radius of identifiability as follows
$$
\fR(\theta) = \sup \{r>0 | \forall \tea \ne \tee, (\|\tea - \theta\| < r, \|\tee - \theta\| < r) \Rightarrow (p_\tea \ne p_\tee)\}
$$
In other words, $\fR(\theta)$ is the largest number, such that the open ball of radius $\fR(\theta)$ around $\theta$ intersected with $\Theta$ 
is an identifiable (sub)family  of probability distributions. If no such ball exists, $\fR(\theta) = 0$.
\end{defn}

From Theorem~\ref{thm:existence_of_poly_algo} and the definition of the radius of identifiability we have the following
\bcor \label{cor:general_case_0}
There exists an algorithm, such that, given  $\epsilon >0$,  for any identifiable $\theta \in \Theta$, where $\Theta$ is the set of parameters within a ball of radius $B$,  it  outputs $\that$ 
within $\min(\epsilon,\fR(\theta))$ of $\te$ with probability $1-\delta$, using a number of  sample points from $p_\te$ polynomial in 
$\max\left(\frac{1}{\epsilon},\frac{1}{\fR(\theta)}\right)$, $\frac{1}{\delta}$ and $B$.
\ecor

\bcor\label{cor:general_case_1}
More generally, if $\theta\in \Theta$, where $\Theta$ is the set of parameters within a ball of radius $B$, is not identifiable but,  ${\cal E}(\theta) = \{\te_1,\ldots,\te_k\}$ is a finite 
set, there exists an algorithm, such that, given  $\epsilon >0$,   it  outputs $\that$ 
within $\min(\epsilon, \min_j \fR(\theta_j))$ of $\te_i$ for some $i \in \{1,\ldots,k\}$ with probability $1-\delta$, using a number of  sample 
points from $p_\te$ polynomial in 
$\max\left(\frac{1}{\epsilon},\frac{1}{\min_j \fR(\theta_j)}\right)$ and $\frac{1}{\delta}$.
\ecor
This last result is what we need to analyze Gaussian mixture model in the next Section.

{\bf Remark:} It is important to note that the radius of identifiability depends on the choice of family $\Theta$. Specifically, the radius is a decreasing function on the family of the sets $\Theta$ ordered by inclusion.

\section{Gaussian Distributions and Polynomially Reducible High Dimensional Families}\label{sec:poly_reducible}

The main result of this section is to show that there exists an algorithm for estimating parameters of  
high-dimensional Gaussian  mixture distributions in time polynomial in the dimension $n$ and other parameters. 
We note that the techniques 
from the previous section cannot be applied directly to high-dimensional  distributions since the 
number of parameters generally increases with dimension. Instead our approach will be to show that 
parameters of high-dimensional  Gaussians can be estimated using $\poly(n)$  linear projections to 
linear subspaces, whose dimension is independent of $n$. We will call this property {\it polynomial reducibility} 
and will also briefly discuss some other families satisfying this condition later in the section.

We will now specifically discuss the case of a mixture of Gaussian distributions.
Let $p_\theta = \sum_{i=1}^k w_i N(\mu_i, \Sigma_i)$ be a mixture of $k$  Gaussian distributions in $\R^n$, with means $\mu_i$ and covariance  matrices $\Sigma_i$.
Let us consider the parameters of the distribution
$\theta = (\mu_1, \Sigma_1,w_1,\ldots,\mu_k, \Sigma_k,w_k)$ as a single vector (thus flattening the covariance matrices). We take the usual Euclidean distance in this space (which, in fact,  corresponds to the Frobenius distance for the covariance matrices).

We will assume that the number of components $k$ is fixed. We note that any permutation of the mixture 
components  leads to the same density function and hence cannot be identified from data. On the other hand, it is 
well known  (\cite{teicher63}) that the density of the distribution determines the parameters uniquely up to a 
permutation, if and only if any two 
components with the same means have different covariance matrices and no mixing coefficient is equal to zero.

The main result of the section is given by the following
\bt\label{thm:gaussian_learning}
Let $p_{\theta}=\sum_{i=1}^kw_i N(\mu_i,\Sigma_i), \theta\in\Theta$, where $\Theta$ is the set of parameters within a ball of radius $B$, be a mixture of Gaussian distributions in $\R^n$ with radius of identifiability $\fR(\theta)$.  Then there exists an algorithm , which, given $\epsilon>0$ and $1>\delta>0$, and $poly\left(n,\max\left(\frac{1}{\epsilon},\frac{1}{\fR(\theta)}\right),\frac{1}{\delta}, B\right)$ samples from $p_{\theta}$,  with probability greater than $(1-\delta)$, outputs a parameter vector  $\hat{\theta}=\left((\hat{\mu}_1, \hat{\Sigma}_1, \hat{w}_1),\ldots,(\hat{\mu}_k, \hat{\Sigma}_k, \hat{w}_k)\right) \in \Theta$,  such that 
there exists a permutation $\sigma:\{1, 2, \ldots, k\}\rightarrow \{1, 2, \ldots, k\}$ satisfying,
$$
\sum_{i=1}^k\left(\|\mu_i-\hat{\mu}_{\sigma(i)}\|^2+\|\Sigma_i-\hat{\Sigma}_{\sigma(i)}\|^2+|w_i-\hat{w}_{\sigma(i)}|^2\right)\leq\epsilon^2
$$
\et
We note that the  radius of identifiability $\fR(\theta)$ can be calculated explicitly from the Proposition~\ref{prop:radius_ident}:
$$
(\fR(\theta))^2 = \min \left(\frac{1}{4} \min_{i \ne j} \left( \|\mu_i - \mu_j\|^2 + \|\Sigma_i - \Sigma_j\|^2 \right), \min_i w_i^2\right)
$$

Thus if the mean/variance pairs for any two components are different with difference bounded from below and 
the minimum mixing weight is is also bounded from below, then we have explicit lower bound for $\fR(\theta)$.

In fact even when $\fR(\theta)$ is not known in advance, it can be estimated from data as:

\bt\label{thm:identifiability_detection}
Let $p_{\theta}=\sum_{i=1}^kw_i N(\mu_i,\Sigma_i), \theta\in\Theta$, where $\Theta$ is the set of parameters within in a ball of radius $B$, be a mixture of Gaussian distributions in $\R^n$ with radius of identifiability $\fR(\theta)$. Then there exists an algorithm , which, given $\epsilon>0$ and $1>\delta>0$, and $poly\left(n,\frac{1}{\epsilon},\frac{1}{\delta}, B\right)$ samples from $p_{\theta}$ outputs whether $\fR(\theta)<\epsilon$ with probability greater than $1-\delta$.
\et

\ep

The rest of the section is structured as follows:

In subsection~\ref{sec:gaussian_properties} we discuss various properties of Gaussian mixture distributions. In particular we derive the 
formula for the radius of identifiability (Proposition~\ref{prop:radius_ident}) and show that there exists a 
low-dimensional projection such that the radius of identifiability changes by at most a linear factor (Theorem~\ref{thm_identifiability}).

In subsection~\ref{sec:sketch_proof} we give a sketch for the proof of the main theorem, showing how the parameters of a high-dimensional distribution can be estimated from a polynomial number of projections. The details of the proof as well as the proof of Theorem~\ref{thm:identifiability_detection} are given in the appendix~\ref{sec:proof}.

Finally, we note  that our results apply to high-dimensional distributions which are not mixtures of Gaussians with a fixed number of components.
For example, a product of $n$ $1$-dimensional Gaussian mixture distributions, which  is a  Gaussian mixture distribution in $n$ dimensions with $k^n$ components,  can be easily learned using our methods. The same applies to other product distributions whose components are polynomial families.





\subsection{Gaussian Distributions}\label{sec:gaussian_properties}
%


\begin{proposition}\label{prop:radius_ident} Let $p_\theta = \sum_{i=1}^k w_i N(\mu_i, \Sigma_i)$, $\theta \in \Theta$ be a family of mixtures of Gaussian distributions in $\R^n$ with non-zero mixing weights.
Then the following inequality is satisfied:
\begin{equation}
\label{eq:radius_identifiability2}
(\fR(\theta))^2 \ge \min \left(\frac{1}{4} \min_{i \ne j} \left( \|\mu_i - \mu_j\|^2 + \|\Sigma_i - \Sigma_j\|^2 \right), \min_i w_i^2\right).
\end{equation}

Moreover, suppose $\Theta$ is a convex set\footnote{Note that requiring convexity is natural, since the set of positive definite matrices is a 
convex cone.} such that it contains all possible mixing coefficients $(w_1,\ldots,w_k)$ for any fixed set of means and variances\footnote{This requirement is unnecessarily strong, however the precise condition, evident from the proof, is awkward to state.} 

In this case the inequality becomes an equality:

\begin{equation}\label{eq:radius_identifiability}
(\fR(\theta))^2 = \min \left(\frac{1}{4} \min_{i \ne j} \left( \|\mu_i - \mu_j\|^2 + \|\Sigma_i - \Sigma_j\|^2 \right), \min_i w_i^2\right)
\end{equation}
In particular, the radius of identifiability is invariant under the permutation of components.

\end{proposition}

\bp We will start by proving the inequality~\ref{eq:radius_identifiability2}.
Suppose  that the distributions $p_{\theta'}$ and $p_{\theta''}$  have the same density. To prove the 
inequality,  we  need to show that at least one of $\te'$, $\te''$ is no closer to $\te$ then the 
right hand side of the inequality~\ref{eq:radius_identifiability2}.

Let us first  consider the case when there is no pair $i\ne j$, s.t. 
$\mu'_i = \mu^{''}_j$ and $\Sigma_i' = \Sigma^{''}_j$. In that case
that  case at  least one of the mixing coefficients for one of the mixtures must be 
equal to zero. That implies that either $\|\te - \te'\|  \ge \min_i w_i $ or $\|\te - \te''\|\ge \min_i  w_i$, which is consistent with the~\ref{eq:radius_identifiability2}.

Alternatively, suppose that for some $i\ne j$ we have $(\mu_i', \Sigma_i') = (\mu_j'', \Sigma_j'')$.
Put $v' = (\mu_i', \Sigma_i') =  (\mu_j'', \Sigma_j'')$, $v_1 = (\mu_i, \Sigma_i)$, $v_2 = (\mu_j, \Sigma_j)$.
We see that 
$$
\|\theta'' -\theta\|^2+\|\theta' -\theta\|^2 \ge 
\|v' - v_1\|^2 + \|v' - v_2\|^2
\ge \frac{1}{2} \|v_1 - v_2\|^2 =
$$
$$
 = \frac{1}{2} \|\mu_i - \mu_j\|^2 +  \frac{1}{2} \|\Sigma_i - \Sigma_j\|^2 
$$
Therefore, $\max\{\|\theta' -\theta\|^2, \|\theta'' -\theta\|^2\}\geq\frac{1}{4} (\|\mu_i - \mu_j\|^2 + \|\Sigma_i - \Sigma_j\|^2)$ which is again  consistent with Inequality~\ref{eq:radius_identifiability2} and together with the first case implies the inequality.

To show Eq.~\ref{eq:radius_identifiability} we need to observe that the bound is tight.
Again we consider two possible cases. If the minimum in the right hand side of 
Eq.~\ref{eq:radius_identifiability} is equal to the square of one of the mixing weights, say, $w_i$, construct $\te'$ 
by putting $w_i'=0$ and keeping the rest of the parameters of $\theta$. We see that $\|\te' - \te\| = w_i$. By slightly perturbing $\mu'$, we see that there exists a $\te''$ arbitrarily close (but not equal)to $\te'$ with the same  probability density. Thus the radius of identifiability cannot exceed $w_i$. 

Alternatively the minimum in the right hand side of Eq.~\ref{eq:radius_identifiability} 
could be equal to $\frac{1}{4}\left(\|\mu_i - \mu_j\|^2 +   \|\Sigma_i - \Sigma_j\|^2\right)$ for some $i \ne j$. Construct $\te'$ by putting $\mu'_i = \mu'_j = \frac{1}{2} (\mu_i - \mu_j)$ and
$\Sigma'_i = \Sigma'_j = \frac{1}{2} (\Sigma_i - \Sigma_j)$ and keeping the rest of the parameters of $\te$. It is easy to see that $\|\te' - \te \|^2 =  \frac{1}{4}\left(\|\mu_i - \mu_j\|^2 +   \|\Sigma_i - \Sigma_j\|^2\right)$. Note that $\theta' \in \Theta$ by the convexity condition. By perturbing $w_i$ and $w_j$ slightly, and keeping the rest of parameters fixed, we can obtain $\te''$ arbitrarily close to $\te'$ with the same probability density.
Hence the radius of identifiability  does not exceed $\frac{1}{4}\left(\|\mu_i - \mu_j\|^2 +   \|\Sigma_i - \Sigma_j\|^2\right)$, which completes the proof.\ep

From the discussion above we have the following 
\begin{cor}\label{cor:identifiability}  Let $\Theta$ be a convex set, such that for any $\theta \in \Theta$ all mixing coefficients $w_i$ are nonzero. Then
\begin{equation}\label{eq:radius_identifiability3}
(\fR(\theta))^2 = \frac{1}{4} \min_{i \ne j} \left( \|\mu_i - \mu_j\|^2 + \|\Sigma_i - \Sigma_j\|^2 \right)
\end{equation}
\end{cor}

It is also easy to see that the radius of identifiability satisfies a type of triangle inequality and that
under any permutation of (mean, covariance matrix, mixing weight) triples the radius of identifiability does not 
change. This is expressed in the following two lemmas (the straightforward proofs are omitted):
\bl\label{lem_radius_of_identifiability}
Let $p_\theta = \sum_{i=1}^k w_i N(\mu_i, \Sigma_i)$, $\theta \in \Theta$ be a family of mixtures of Gaussian distributions in $\R^n$. For any $\theta_1, \theta_2\in\Theta$ such that $\theta_1\in\mathcal{N}(\theta_2,\epsilon)$ for some $\epsilon>0, |\fR(\theta_1)-\fR(\theta_2)|\leq\epsilon$.
\el

\bl
Let $p_{\theta}=\sum_{i=1}^kw_i N(\mu_i,\Sigma_i), \theta\in\Theta$ be a mixture of Gaussian distributions in $\R^n$. Suppose $\theta$ is represented as $\theta=(\theta_1,\theta_2,\ldots,\theta_k)$, where $\theta_i=(\mu_i,\Sigma_i,w_i)$ is the mean, covariance matrix, mixing weight triple.
Let $\theta^{'}=(\theta_{\sigma(1)}, \theta_{\sigma(2)},\ldots,\theta_{\sigma(2)})$, 
where $\sigma :\{1,2,\ldots,k\}\rightarrow \{1,2,\ldots,k\}$ is a permutation. Then $\fR(\theta)=\fR(\theta^{'})$.
\el

%

From now on, we will assume that $\Theta$ is a sufficiently large ball or cube (with the necessary conditions to 
make $p_\te$ a valid probability distribution), so that we do not have to worry about 
convexity and other technical properties. 

We now recall that a projection of a Gaussian mixture distribution onto a subspace is a lower-dimensional Gaussian mixture distribution. 
Specifically, if 
$p_{\theta}=\sum_{i=1}^k w_i N(\mu_i, \Sigma_i)$,  the Gaussian mixture distributions in $\R^n$, is projected onto a subspace $S$ then the projection is a lower-dimensional Gaussian mixture distribution family $\pi_S(p_{\theta})$, parameterized by $P_S(\theta)$.  
In particular, if $S$ is a coordinate plane then $P_S$ is a projection operator, which is an identity mapping for 
the mixing weights, an orthogonal projection onto $S$ for the means and 
the restriction operator for the covariance matrices of the components, where each covariance matrix
is projected to its minor corresponding to the coordinates in $S$.


 We will now state the following Theorem whose proof can be found in appendix~\ref{sec:proof}.
\bt\label{thm_identifiability}
Let $p_{\theta}=\sum_{i=1}^kw_iN(\mu_i,\Sigma_i)$, $\theta \in \Theta$ be a Gaussian mixture distribution in $\R^n$ with  radius of identifiability  $\fR(\theta)$.
Then there exists a $2k^2$-dimensional coordinate plane $S$, such that $\fR(P_S(\theta))\geq\frac{1}{n}\fR(\theta)$.
\et

\subsection{Sketch of the Proof of Theorem~\ref{thm:gaussian_learning}}\label{sec:sketch_proof}
We present a brief overview of the proof. The technical details can be found in Appendix~\ref{sec:proof}. 
The main idea is to show that 
parameters of high-dimensional Gaussian mixture can be estimated arbitrarily well using $\poly(n)$  projections to 
coordinate subspaces, whose dimension only depends on $k$. Since the dimension of these lower dimensional subspaces is independent of $n$, results from Section~\ref{sec:poly_families} can be used to estimate the parameters.

Let $\theta=(\theta_1,\theta_2,\ldots,\theta_k)$, where $\theta_i=(\mu_i,\Sigma_i,w_i)$, be the parameter vector after flattening the covariance matrices. Recall that projection of $p_\theta$ onto a $2k^2$-coordinate plane $T$, will result in a mixture $\pi_T(p_\theta)$, parameterized (with a slight abuse of notation) by $P_T(\theta)=(P_T(\theta_1),P_T(\theta_2),\ldots,P_T(\theta_k))$.

\noindent\textbf{Step 1:} 
Let $\fR(\theta)$ be the radius of identifiability.  
Theorem~\ref{thm_identifiability} guarantees the existence of a $2k^2$-dimensional coordinate subspace $S$, such that radius of identifiability decreases by at most $\frac{1}{n}$, $\fR(P_S(\theta))\geq\frac{1}{n}\fR(\theta)$. 

To identify such a subspace, we take all $ n\choose 2k^2$ coordinate projections. For each projection to a subspace 
$T$ we estimate  the parameters using the Theorem~\ref{thm:existence_of_poly_algo}. It is important to note that 
given  $\epsilon'$ as input, Theorem~\ref{thm:existence_of_poly_algo} is guaranteed to produce a value of parameter 
$\widehat{P_T(\theta)}$, such that $|\fR(\widehat{P_T(\theta)}) - \fR(P_T(\te))| < \epsilon'$ (Lemma~\ref{lem_radius_of_identifiability}) using a number of samples polynomial in $k$ and $\frac{1}{\epsilon'}$. 
Applying the union bound for all $n\choose 2k^2$ projections provides an estimate for the radius of identifiability 
for each projection within $\epsilon'$. Choosing $\epsilon'$ appropriately (say, $\frac{\fR(\te)}{2n}$), and 
choosing the projection with the largest estimated radius of identifiability, yields a coordinate subspace $S$ with 
a  lower bounded $\fR(P_S(\te))$.

\begin{wrapfigure}{l}{0.5\textwidth}
\centerline{\includegraphics[scale=0.35]{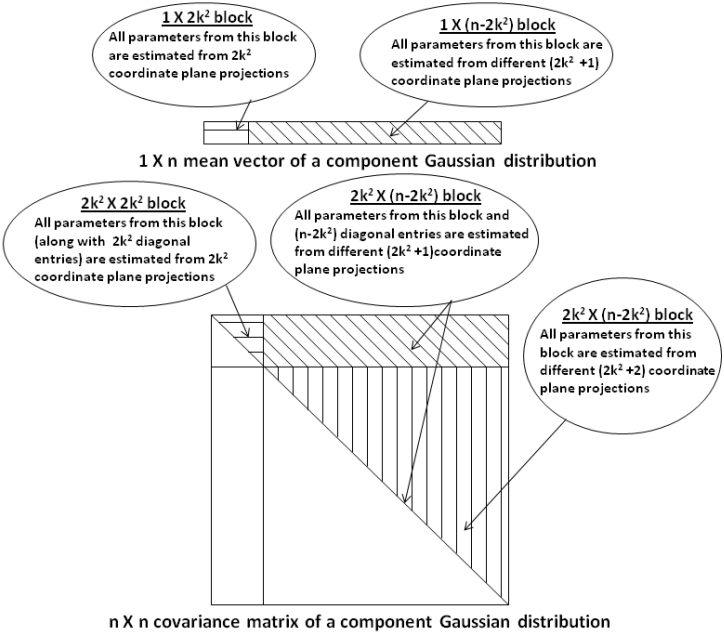}}
\caption{\small Estimation of high-dimensional Gaussian mixture parameters from $\poly(n)$ lower-dimensional projections.}
\label{fig:reconstruction}
\vspace{-0.25in}
\end{wrapfigure}

The coordinates within $S$ are represented by the horizontally shaded region in  Figure~\ref{fig:reconstruction}.
We use this space as a starting point for Step~2.

\noindent\textbf{Step 2:} By applying Corollary~\ref{cor:general_case_1} to the projection $P_S(\te)$,
we can estimate the mixing weights, projections of the original means and $2k^2 \times 2k^2$ minors of the 
covariance matrices corresponding to the coordinates within $S$. We now need to estimate the rest of the parameters 
using a sample size polynomial on $n$. We do this by estimating each additional coordinate separately. 
That is for each coordinate $i$ not in $S$ we take $S_i = span(S, e_i)$, where $e_i$ is the corresponding coordinate vector. It can be seen that the radius of 
identifiability does not decrease going from $S$ to $S_i$. We show that the $i$'th coordinate of each component 
mean can be estimated by applying Corollary~\ref{cor:general_case_1} to the projection to $S_i$. We repeat this 
procedure for each of the $n-2k^2$ coordinates not in $S$. 

To estimate the covariance matrices we proceed similarly, except that we need to estimate entries 
corresponding to pairs of coordinates $(i,j)$. Now we have two possibilities, since either one of $i,j$ or both of 
them may not be in $S$. If exactly one of them, say $i$, is not in $S$, projection to $S_i$ defined above can be 
used to estimate the corresponding entry of each covariance matrix. If both $i,j$ are not in $S$, we take the projection onto $S_{ij} = span(S, e_i, e_j)$. By applying Corollary~\ref{cor:general_case_1}, we show that the $ij$'th entry of covariance matrices can also be estimated.  

Thus, after obtaining the initial space $S$,  the complete set of parameters can be estimated using at most $n-2k^2 + {{n-2k^2}\choose 2}$ parameter estimations for  $2k^2+1$ or $2k^2+2$-dimensional subspaces.

This procedure is graphically shown in in Figure~\ref{fig:reconstruction}.

\section{Conclusion and Discussion}\label{sec:conclusion}

The results of this paper resolve the general problem of polynomial learning of Gaussian mixture 
distributions.  Our results do not require any separation assumptions and apply 
as long as the mixture is identifiable. For example, they apply even if all components of the 
mixture have the same mean distribution, as long as the covariance matrices are different and the mixing coefficients are non-zero.

The proof brings the techniques of algebraic geometry to the classical method of moments, an approach that, as far 
as we know, is new to this domain. We also provide 
quite general results applicable to learning various low-dimensional families and some  
observations on high-dimensional families going  beyond Gaussian mixture distributions with a fixed number of 
components. For example, one can also learn products of arbitrary probability distributions in a fixed  low-dimensional  polynomial family, e.g., a product of $n$ number of $d$-dimensional Gaussians mixtures with $k$ components each (which 
is a $nd$-dimensional Gaussian mixture distribution with $k^n$ components).

We are planning to investigate other applications in learning of the framework presented in this paper.
We also note that the methods proposed in the paper can be turned into
implementable (and potentially practical) algorithms through the use of tools from computational algebraic geometry.
This is also a direction of future investigation.

%

\bibliographystyle{plain}
\bibliography{current_work}

\appendix
\section*{Appendix}

\section{Some Polynomial Families of Distributions}\label{app:distribution_family}
In this appendix we provide a partial list comprising the expressions of moments of various univariate probability distributions which form polynomial families. It turns out that most of the commonly used distributions 
form polynomial families as shown Table \ref{tab:poly_table}. 
In the fifth column of Table \ref{tab:poly_table}, we provide either expression for the $i^{th}$ moment or a recurrence relation, which shows that the moments are polynomial in the distribution parameters, along with explicit expressions for the first three moments. These moment expressions and recurrence relations are well known and can be found in, e.g., \cite{feller71, riordan37}. In a couple of cases we need a slightly different parameterization, instead of the standard one, to ensure that the moments are polynomial in these new parameters. For example, in standard parametrization, Negative Binomial distribution $NB(r,p)$ is expressed by probability mass function ${x+r-1\choose r-1}(1-p)^r p^x$. However, if we  replace $p$ by a new parameter $m=\frac{p}{1-p}$, then the moments are polynomial in $r$ and $m$. Recurrence relation for this new parameterization can be obtained following the same steps as in~\cite{riordan37}. Table \ref{tab:non_polynomial} we list two families which are not polynomial.

\begin{table}
\centering
\begin{tabular}{| @{\hspace{.5mm}}c@{\hspace{.5mm}} || @{\hspace{.5mm}}c@{\hspace{.5mm}} | @{\hspace{.1mm}}c@{\hspace{.1mm}} | @{\hspace{.1mm}}c@{\hspace{.1mm}} |l| }\hline
Distribution & $\theta$ & Pdf/Pmf  $f(x;\theta)$& Mgf $M(t)$  &~~~~~~~~~~~~~~~~Moments Expression\\ \hline

\multirow{4}{*}{Gaussian} & \multirow{4}{*}{$\mu,\sigma$} &    
\multirow{4}{*}{$\frac{1}{\sqrt{2\pi}\sigma}e^{-\frac{(x-\mu)^2}{2\sigma^2}}$} & \multirow{4}{*}{$e^{\mu t+\frac{\sigma^2t^2}{2}}$} &
\multicolumn{1}{>{\columncolor[rgb]{.8,.8,.8}}c|}{$\mathbb{E}(X^i)=\mu \mathbb{E}(X^{i-1})+(i-1)\sigma^2 \mathbb{E}(X^{i-2})$}\\ 
\cline{5-5}
 & & & & $\mathbb{E}(X)~=\mu$\\  
 & & & & $\mathbb{E}(X^2)=\mu^2+\sigma^2$\\
 & & &  &$\mathbb{E}(X^3)=\mu^3+3\mu\sigma^2$\\ \hline

\multirow{4}{*}{Uniform} & \multirow{4}{*}{$a,b$} & \multirow{4}{*}{$
\frac{1}{b-a} \mbox{, $a\leq x\leq b$}
$} &\multirow{4}{*}{$\frac{e^{tb}-e^{ta}}{t(b-a)}$} & 
\multicolumn{1}{>{\columncolor[rgb]{.8,.8,.8}}c|}{$\mathbb{E}(X^i)=\frac{1}{i+1}\sum_{j=1}^i a^jb^{i-j}$}\\ 
\cline{5-5}
& & & & $\mathbb{E}(X)~=\frac{a+b}{2}$\\ 
& && &  $\mathbb{E}(X^2)=\frac{a^2+ab+b^2}{3}$\\
& & & & $\mathbb{E}(X^3)=\frac{a^3+a^b+ab^2+b^3}{4}$\\ \hline

\multirow{4}{*}{Gamma} & \multirow{4}{*}{$\beta, m$} & \multirow{4}{*}{$
\frac{x^{m-1}e^{-x/\beta}}{\beta^m\Gamma(m)} \mbox{, $x>0$} 
$} & \multirow{4}{*}{$(1-\beta t)^{-m}$} & 
\multicolumn{1}{>{\columncolor[rgb]{.8,.8,.8}}c|}{$\mathbb{E}(X^i)=\prod_{j=0}^{i-1}(m+j)\beta^i$}\\  \cline{5-5}
& & & & $\mathbb{E}(X)~=m\beta$\\ 
& & & & $\mathbb{E}(X^2)=m(m+1)\beta^2$\\
& & & & $\mathbb{E}(X^3)=m(m+1)(m+2)\beta^3$\\ \hline

\multirow{4}{*}{Laplace} & \multirow{4}{*}{$\mu, b$} & \multirow{4}{*}{$
\frac{1}{2b}e^{-\frac{|x-\mu|}{b}} 
$} &\multirow{4}{*}{$\frac{e^{\mu t}}{1-b^2t^2}$} & 
\multicolumn{1}{>{\columncolor[rgb]{.8,.8,.8}}c|}{$\mathbb{E}(X^i)=\sum_{j=0}^{i}\frac{i!b^j\mu^{i-j}}{(i-j)!}1_{\{j \mbox{ is even}\}}$}\\ 
\cline{5-5}
& & & & $\mathbb{E}(X)~=\mu$\\ 
& & & & $\mathbb{E}(X^2)=\mu^2+2b^2$\\
& & & & $\mathbb{E}(X^3)=\mu^3+6\mu b^2$\\ \hline

\multirow{4}{*}{Exponential} & \multirow{4}{*}{$\lambda$} & \multirow{4}{*}{$
\frac{1}{\lambda}e^{-\frac{x}{\lambda}} \mbox{, $x>0$} 
$} & \multirow{4}{*}{$(1-\lambda t)^{-1}$} & 
\multicolumn{1}{|>{\columncolor[rgb]{.8,.8,.8}}c|}{$\mathbb{E}(X^i)=i!\lambda^i$}\\  \cline{5-5}
& & & & $\mathbb{E}(X)~=\lambda$\\ 
& & & & $\mathbb{E}(X^2)=2\lambda^2$\\
& & & & $\mathbb{E}(X^3)=6\lambda^3$\\ \hline

\multirow{4}{*}{Chi-Square} & \multirow{4}{*}{$k$} & \multirow{4}{*}{$
\frac{x^{\frac{k}{2}-1}e^{-\frac{x}{2}}}{2^{k/2}\Gamma(k/2)} \mbox{, $x>0$} 
$} & \multirow{4}{*}{$(1-2t)^{-\frac{k}{2}}$} &  
\multicolumn{1}{>{\columncolor[rgb]{.8,.8,.8}}c|}{$\mathbb{E}(X^i)=k(k+2)\cdots(k+2i-2)$}\\ \cline{5-5}
& & & & $\mathbb{E}(X)~=k$\\ 
& & & & $\mathbb{E}(X^2)=k(k+2)$\\
& & & & $\mathbb{E}(X^3)=k(k+2)(k+4)$\\ \hline

 & \multirow{4}{*}{$\mu, \lambda$} & \multirow{4}{*}{$
\sqrt{\frac{1}{2\pi\lambda x^3}}e^{-\frac{(x-\mu)^2}{2\lambda\mu^2x}} 
$} & \multirow{4}{*}{$e^{\frac{(1-\sqrt{1-2\lambda\mu^2t})}{\lambda\mu}}$} &  
\multicolumn{1}{>{\columncolor[rgb]{.8,.8,.8}}c|}{$\mathbb{E}(X^i)=(2i-3)\lambda\mu^2\mathbb{E}(X^{i-1})+\mu^2\mathbb{E}(X^{i-2})$}\\ \cline{5-5}
Inverse & & & & $\mathbb{E}(X)~=\mu$\\  
Gaussian & & & & $\mathbb{E}(X^2)=\lambda\mu^3$\\
& & & & $\mathbb{E}(X^3)=3\lambda^2\mu^5$\\ \hline

\multirow{4}{*}{Poisson} & \multirow{4}{*}{$\lambda$} & \multirow{4}{*}{$
\frac{\lambda^x e^{-\lambda}}{x!}
$} & \multirow{4}{*}{$e^{\lambda(e^t-1)}$} &  
\multicolumn{1}{>{\columncolor[rgb]{.8,.8,.8}}c|}{$\mathbb{E}(X^i)=\lambda \mathbb{E}(X^{i-1})+\lambda \frac{d(\mathbb{E}(X^{i-1}))}{d\lambda}$}\\ \cline{5-5}
& & & & $\mathbb{E}(X)~=\lambda$\\ 
& & & & $\mathbb{E}(X^2)=\lambda^2+\lambda$\\
& & & & $\mathbb{E}(X^3)=\lambda^3+3\lambda^2+\lambda$\\ \hline

\multirow{4}{*}{Binomial} & \multirow{4}{*}{$n, p$} & \multirow{4}{*}{$
{n\choose x}p^x(1-p)^{n-x} 
$} & \multirow{4}{*}{$(1-p+pe^t)^n$} &  
\multicolumn{1}{>{\columncolor[rgb]{.8,.8,.8}}c|}{$\mathbb{E}(X^i)=np\mathbb{E}(X^{i-1})+p(1-p)\frac{d(\mathbb{E}(X^{i-1}))}{dp}$}\\ \cline{5-5}
& & & & $\mathbb{E}(X)~=np$\\  
& & & & $\mathbb{E}(X^2)=n(n-1)p^2+np$\\
& & & & $\mathbb{E}(X^3)=(n^3-3n^2+2n)p^3+3n(n-1)p^2+np$\\ \hline

\multirow{4}{*}{Geometric} & \multirow{4}{*}{$p$} & \multirow{4}{*}{$
(1-\frac{1}{p})^x(\frac{1}{p}) 
$} & \multirow{4}{*}{$\frac{1}{p-(p-1)e^t}$} &  
\multicolumn{1}{>{\columncolor[rgb]{.8,.8,.8}}c|}{$\mathbb{E}(X^i)=\sum_{j=0}^{\infty}\frac{1}{p}\left(1-\frac{1}{p}\right)^j j^i$}\\ \cline{5-5}
& & & & $\mathbb{E}(X)~=(p-1)$\\ 
& & & & $\mathbb{E}(X^2)=(p-1)(2p-1)$\\ 
& & & & $\mathbb{E}(X^3)=(p-1)(6p^2-6p+1)$\\ \hline

 & \multirow{4}{*}{$r, m$} & \multirow{4}{*}{$
{x+r-1\choose r-1}\frac{m^x}{(m+1)^{r+x}} 
$} & \multirow{4}{*}{$\left(\frac{1}{m+1-me^t}\right)^r$} &  
\multicolumn{1}{>{\columncolor[rgb]{.8,.8,.8}}c|}{$\mathbb{E}(X^i)=rm\mathbb{E}(X^{i-1})+m(m+1)\frac{d(\mathbb{E}(X^{i-1}))}{dm}$}\\ \cline{5-5}
Negative & & & & $\mathbb{E}(X)~=rm$\\  
Binomial & & & & $\mathbb{E}(X^2)=r(r+1)m^2+rm$\\
& & & & $\mathbb{E}(X^3)=(r^3+3r^2+2r)m^3+3r(r+1)m^2+rm$\\ \hline
\end{tabular}
\caption{Common polynomial families and their moments}
\label{tab:poly_table}
\end{table}

\begin{table}[]
\centering
\begin{tabular}{| @{\hspace{.5mm}}c@{\hspace{.5mm}} || @{\hspace{.5mm}}c@{\hspace{.5mm}} | @{\hspace{.1mm}}c@{\hspace{.1mm}} | @{\hspace{.1mm}}c@{\hspace{.1mm}} |l| }\hline
Distribution & $\theta$ & Pdf/Pmf  $f(x;\theta)$& Mgf $M(t)$  &Moments Expression\\ \hline
Weibull & $k, \lambda$ & $\frac{k}{\lambda}\left(\frac{x}{\lambda}\right)^{k-1}e^{-\left(\frac{x}{\lambda}\right)^k}$ & $\sum_{n-0}^{\infty}\frac{t^n\lambda^n}{n!}\Gamma\left(1+\frac{n}{k}\right)$ & $\mathbb{E}(X^i)=\lambda^i\Gamma\left(1+\frac{i}{k}\right)$ \\ \hline
Cauchy & $\lambda, \gamma$ & $\frac{1}{\pi\gamma\left[1+\left(\frac{x-\lambda}{\gamma}\right)^2\right]}$ & Does not exist & Does not exist \\ \hline
\end{tabular}
\caption{Examples of some probability distributions that do not belong to polynomial family}
\label{tab:non_polynomial}
\end{table}

\section{Separation Preserving Coordinate Planes}
Let $p_\theta = \sum_{i=1}^k w_i N(\mu_i, \Sigma_i)$ be a mixture of $k$  Gaussian distributions in $\R^l$, with means $\mu_i$ and covariance  matrices $\Sigma_i$. When this distribution is projected onto any lower dimensional coordinate plane $S$, the corresponding Gaussian mixture $\pi_S(p_\theta)$, parameterized by $P_S(\theta)$, has means and covariance matrices represented by $P_S(\mu_i)$ and $P_S(\Sigma_i)$ respectively.
We first show that if any pair of means or pair of covariance matrices of the
original component Gaussian distributions are separated, then they remain so after
projecting the mixture distribution onto some suitable lower dimensional coordinate plane.

\noindent\textbf{Existence of a Coordinate Plane where Projected Means Remain Separated :}
\begin{lemma}\label{lem:unique_mean}
For any $\mu_1, \mu_2,...,\mu_k \in \mathbb{R}^l$, there exists a $k^2$-coordinate plane $S$ such that, 
$$\forall_{i,j}, \|P_{S}(\mu_i)-P_{S}(\mu_j)\|\geq\|\mu_i-\mu_j\|\frac{1}{\sqrt{l}}$$.
\end{lemma}
\bp
We will use $\mathcal{M}$ to denote a set of indices of coordinate directions of $\R^l$ and let $S_{\mathcal{M}}$ be the $|\mathcal{M}|$-coordinate plane, where $|\mathcal{M}|$ is the cardinality of $\mathcal{M}$, spanned by the coordinate directions whose indices are in $\mathcal{M}$. Initially $\mathcal{M}$ is empty.
Let $\mathcal{A}_i=\left\{1,2,\ldots,i-1,i+1,\ldots,k\right\}$.
Now consider the pair consisting
of $\mu_1$ and any other $\mu_j$ such that $j\in\mathcal{A}_1$.
There exists at least one coordinate direction, whose index is say $m$, such that
$|\mu_{1,m}-\mu_{j,m}|\geq\|\mu_1-\mu_j\|\frac{1}{\sqrt{l}}$.
Adding $m$ to $\mathcal{M}$, and projecting onto $S_{\mathcal{M}}$, guarantees that $\|P_{S_{\mathcal{M}}}(\mu_1)-P_{S_{\mathcal{M}}}(\mu_j)\|\geq\|\mu_1-\mu_j\|\frac{1}{\sqrt{l}}$.
Note that for
$\mu_1$, in the worst case, we may have to include indices of $(k-1)$
extra coordinate directions to $\mathcal{M}$ to ensure that after projection onto
$S_{\mathcal{M}},  \|P_{S_{\mathcal{M}}}(\mu_1)-P_{S_{\mathcal{M}}}(\mu_j)\|\geq\|\mu_1-\mu_j\|\frac{1}{\sqrt{l}}$
for any $j\in\mathcal{A}_1$.

Similarly, in addition, to ensure that that after projecting onto  $S_{\mathcal{M}}$, $P_{S_{\mathcal{M}}}(\mu_2)$
is guaranteed to remain separated from any $P_{S_{\mathcal{M}}}(\mu_j)$, $j\in\mathcal{A}_2$, by at least
$\|\mu_2-\mu_j\|\frac{1}{\sqrt{l}}$, we may need to add indices of $(k-2)$ additional coordinate directions to $\mathcal{M}$ and so on. So in total $\mathcal{M}$ can have indices of at most $(k-1)+(k-2)+\cdots+1=\frac{k(k-1)}{2}<k^2$ coordinate directions to ensure that as long as we
project $\{\mu_i\}_{i=1}^k$  onto $l \choose k^2$ different $k^2$-coordinate planes, there exists at least
one $k^2$-coordinate plane $S$ 
such that $\forall_{i,j},~ \|P_{S}(\mu_i)-P_{S}(\mu_j)\|\geq\|\mu_i-\mu_j\|\frac{1}{\sqrt{l}}$.
\ep


\noindent\textbf{Existence of a Coordinate Plane where Projected Covariance Matrices Remain Separated :}
\begin{lemma}\label{lem:unique_var}
For any $\Sigma_1, \Sigma_2,...,\Sigma_k \in \R^{l\times l}$, there exists a $k^2$-coordinate plane $S$ such that,
$$\forall_{i,j}, \|P_S(\Sigma_i)-P_S(\Sigma_j)\|\geq\|\Sigma_i-\Sigma_j\|\frac{1}{l}$$.
\end{lemma}
\bp
We will use $\mathcal{M}$ to denote a set of indices of coordinate directions of $\R^l$ and let $S_{\mathcal{M}}$ be the $|\mathcal{M}|$-coordinate plane, where $|\mathcal{M}|$ is the cardinality of $\mathcal{M}$, spanned by the coordinate directions whose indices are in $\mathcal{M}$. Initially $\mathcal{M}$ is empty.
Let $\mathcal{A}_i=\left\{1,2,\ldots,i-1,i+1,\ldots,k\right\}$.
Now consider the pair consisting
of $\Sigma_1$ and any other $\Sigma_j$ such that $j\in\mathcal{A}_1$.
Since $\Sigma_1$ and $\Sigma_j$ must differ in at least one diagonal or off-diagonal
element by an amount $\|\Sigma_1-\Sigma_j\|\frac{1}{l}$, there must exist two coordinate directions, whose indices are say, $p$ and $q$, such that adding $p$ and $q$ to $\mathcal{M}$ and projecting onto $S_{\mathcal{M}}$ guarantees that $\|P_{S_{\mathcal{M}}}(\Sigma_1)-P_{S_{\mathcal{M}}}(\Sigma_j)\|\geq\|\Sigma_1-\Sigma_j\|\frac{1}{l}$.
Note that for
$\Sigma_1$, in the worst case, we may have to add indices of $2(k-1)$
extra coordinate directions to $\mathcal{M}$ to ensure that that after projecting onto 
$S_{\mathcal{M}}, \|P_{S_{\mathcal{M}}}(\Sigma_1)-P_{S_{\mathcal{M}}}(\Sigma_j)\|\geq\|\Sigma_1-\Sigma_j\|\frac{1}{l}$
for any $j\in\mathcal{A}_1$.

Similarly, in addition, to ensure that that after projecting onto $S_{\mathcal{M}}$, $P_{S_{\mathcal{M}}}(\Sigma_2)$
is guaranteed to remain separated from any $P_{S_{\mathcal{M}}}(\Sigma_j)$, $j\in\mathcal{A}_2$, by
at least $\|\Sigma_2-\Sigma_j\|\frac{1}{l}$, we may need to add indices of $2(k-2)$ additional coordinate directions to $S_{\mathcal{M}}$ and so on. So in total $\mathcal{M}$ can have indices of at most $2(k-1)+2(k-2)+\cdots+1=k(k-1)<k^2$ coordinate directions to ensure that as long as we
project $\{\Sigma_i\}_{i=1}^k$  onto $l \choose k^2$ different  $k^2$-coordinate planes, there exists at least
one $k^2$-coordinate plane $S$ 
such that $\forall_{i,j},~ \|P_S(\Sigma_i)-P_S(\Sigma_j)\|\geq\|\Sigma_i-\Sigma_j\|\frac{1}{l}$.
\ep

\section{Proof of Theorem~\ref{thm:gaussian_learning},  Theorem~\ref{thm:identifiability_detection} and Theorem~\ref{thm_identifiability}}\label{sec:proof}
In this appendix we give the detailed proof of Theorem~\ref{thm:gaussian_learning} as well as proof of Theorem~\ref{thm:identifiability_detection} and Theorem~\ref{thm_identifiability}. We start with some preliminary Lemmas.
\bl\label{lem:component_separation}
Let $p_{\theta}=\sum_{i=1}^kw_i N(\mu_i,\Sigma_i)$, where $\theta$ is the set of parameters within a ball of radius $B$, be a mixture of Gaussian distributions in $\R^n$ with the radius of identifiability $\fR(\theta)$. If $\theta$ is represented as $\theta=(\theta_1,\theta_2,\ldots,\theta_k)$, where $\theta_i=(\mu_i,\Sigma_i,w_i)$ is the mean, covariance matrix, mixing weight triple, (after flattening the covariance matrices) then for any $i\neq j, \|\theta_i-\theta_j\|\geq2\fR(\theta)$.
\el

\bp
Explicit expression for $\fR(\theta)$ is given in Equation~\ref{eq:radius_identifiability}. If  $\frac{1}{4} \min_{i \ne j} \left( \|\mu_i - \mu_j\|^2 + \|\Sigma_i - \Sigma_j\|^2 \right)<\min_i w_i^2$ then for any $i\neq j, (\fR(\theta))^2=\frac{1}{4} \min_{i \ne j} \left( \|\mu_i - \mu_j\|^2 + \|\Sigma_i - \Sigma_j\|^2 \right)\leq\frac{1}{4}\|\theta_i-\theta_j\|^2$. On the other hand if $\frac{1}{4} \min_{i \ne j} \left( \|\mu_i - \mu_j\|^2 + \|\Sigma_i - \Sigma_j\|^2 \right)\geq\min_i w_i^2$ then for any $i\neq j$,  $(\fR(\theta))^2=\min_i w_i^2\leq\frac{1}{4} \min_{i \ne j} \left( \|\mu_i - \mu_j\|^2 + \|\Sigma_i - \Sigma_j\|^2 \right)\leq\frac{1}{4}\|\theta_i-\theta_j\|^2$.
\ep

\bl\label{lem:ri_preservation}
Let $p_{\theta}=\sum_{i=1}^kw_i N(\mu_i,\Sigma_i)$, where $\theta$ is the set of parameters within a ball of radius $B$, be a mixture of Gaussian distributions in $\R^n$ with radius of identifiability $\fR(\theta)$. Let $S$ and $T$ be two lower-dimensional subspaces such that $S\subset T$. 
Then $\fR(P_T(\theta))\geq\fR(P_S(\theta))$.
\el
\bp
Immediate from Equation~\ref{eq:radius_identifiability}.
\ep

\noindent\textbf{Proof of Theorem~\ref{thm:gaussian_learning} :}


Let $\theta=(\theta_1,\theta_2,\ldots,\theta_k)$, where $\theta_i=(\mu_i,\Sigma_i,w_i)$,  be parameter vector after flattening the covariance matrices. Recall that projection of $p_\theta$ onto any $2k^2$-coordinate plane $T$, will result in a mixture $\pi_T(p_\theta)$, which is parameterized (with a little abuse of notation) by $P_T(\theta)=(P_T(\theta_1),P_T(\theta_2),\ldots,P_T(\theta_k))$.

\noindent \textbf{Details of Step 1:}  
Let $\gamma=\min\left(\frac{\fR(\theta)}{n},\frac{\epsilon}{n}\right)$ 
where $\fR(\theta)$ is the radius of identifiability.  
Theorem~\ref{thm_identifiability} guarantees the existence of a $2k^2$-dimensional coordinate subspace $S$, such that radius of identifiability decreases by at most $\frac{1}{n}$, $\fR(P_S(\theta))\geq\frac{1}{n}\fR(\theta)\geq\gamma$.
To identify such a subspace, we take all $ n\choose 2k^2$ coordinate projections. 
For any fixed projection to a $2k^2$-dimensional subspace 
$T$, invoking Theorem~\ref{thm:existence_of_poly_algo} using a sample of size $\poly(\frac{1}{\gamma},\frac{1}{\delta},B)$, (setting the precision parameter to $\frac{\gamma}{3}$) produces a value of parameters $\widehat{P_T(\theta)}$ such that 
$|\fR(\widehat{P_T(\theta)}) - \fR(P_T(\te))| < \frac{\gamma}{3}$ (Lemma~\ref{lem_radius_of_identifiability}). 
Applying the union bound for all $n\choose 2k^2$ projections provides an estimate for the radius of identifiability 
for each projection within $\frac{\gamma}{3}$. Thus invoking  Theorem~\ref{thm:existence_of_poly_algo} $n\choose 2k^2$ times, each time using a sample of size $\poly\left(\frac{1}{\gamma},\frac{1}{(\delta/4n^2)},B\right)$,  (setting the precision parameter to $\frac{\gamma}{3}$) and
choosing the projection with the largest estimated radius of identifiability, yields a coordinate subspace $S$ such that with probability at least $1-\frac{\delta}{4},~\fR(P_S(\theta))\geq\frac{\gamma}{3}$. Clearly the sample size requirement for this step is polynomial in $n$.

\noindent \textbf{Details of Step 2:}
By applying Corollary~\ref{cor:general_case_1} to the mixture $\pi_S(p_\theta)$, where $S$ is obtained in Step 1, using a sample of size $\poly(\frac{1}{\gamma},\frac{1}{\delta},B)$, (setting the precision parameter to $\frac{\gamma}{9}$) with probability greater than $1-\frac{\delta}{4}$ we can get an estimate of $\widehat{P_S(\theta)}$ satisfying $\|\widehat{P_S(\theta)}-P_S(\theta)\|\leq\frac{\gamma}{9}$. Note that these estimates encompass the mixing weights, projections of the original means and $2k^2 \times 2k^2$ minors of the covariance matrices corresponding to the coordinates within $S$.
If we let $\theta^{'}$ to be $P_S(\theta)$ then the estimate $\hat{\theta}^{'}=\widehat{P_S(\theta)}$ is, up to a permutation, within $\frac{\gamma}{9}$ of $\theta^{'}$ with probability greater than $(1-\frac{\delta}{4})$.
Note that the dimension of $\theta^{'}$ is $(k-1)+k\left(2k^2+\frac{2k^2(2k^2+1)}{2}\right)$. These parameters are represented by the horizontally shaded region in  Figure \ref{fig:reconstruction}.

We now need to estimate the rest of the parameters 
using a sample size polynomial on $n$.  This procedure explained in the following two sub-steps.

\noindent \textbf{2a: Estimating means and part of covariance matrices}

In this sub-step we estimate each additional coordinate separately.
That is for each coordinate $i$ not in $S$ we take $S_i = span(S, e_i)$, where $e_i$ is the corresponding coordinate vector. It can be seen that the radius of 
identifiability does not decrease going from $S$ to $S_i$. We will show that the $i$'th coordinate of each component 
mean, $i$'th diagonal entry for each component covariance matrix and $2k^2$  extra off diagonal entries for each component covariance matrix can be estimated by applying Corollary~\ref{cor:general_case_1} to the projection to $S_i$. We repeat this 
procedure for each of the $n-2k^2$ coordinates not in $S$.


For each such $n-2k^2$ coordinates not in $S$, we project $p_{\theta}$ onto $S_i$ and invoke the algorithm of 
Corollary~\ref{cor:general_case_1} (setting the precision parameter to be $\frac{\gamma}{9}$) using a sample of size $\poly\left(\frac{1}{\gamma},\frac{1}{(\delta/4n)},B\right)$ . Clearly this sample size is polynomial in $n$.
This ensures that, each time we get an estimate $\widehat{P_{S_i}(\theta)}$ such that with probability at least $1-\frac{\delta}{4},~\|\widehat{P_{S_i}(\theta)}-P_{S_i}(\theta)\|\leq\frac{\gamma}{9}$.
Since 
$P_S(\theta)\subset P_{S_i}(\theta)$ letting $\phi_i=P_{S_i}(\theta)\setminus P_S(\theta)$ to be the extra parameters, we have for each $i$, 
$$\|\hat{\phi}_i-\phi_i\|=\|\widehat{P_{S_i}(\theta)}-P_{S_i}(\theta)\|\leq\frac{\gamma}{9}$$ 
with probability greater than $(1-\frac{\delta}{4})$, where $\hat{\phi}$ is the estimate of $\phi$. 
Since for each $S_i$, $\forall_{m\neq n}, \|P_{S_i}(\theta_m)-P_{S_i}(\theta_n)\|\geq\frac{2\gamma}{3}$, (using Lemma~\ref{lem:component_separation} and Lemma~\ref{lem:ri_preservation}), estimates of the extra parameters can be uniquely associated to the parameters of the component Gaussian distributions estimated in Step 2.

%

Letting $\theta^{''}$ to be $\cup_{i=1}^{n-2k^2}\phi_i$, we have 
$$
\|\hat{\theta}^{''}-\theta^{''}\|=\left(\sqrt{\sum_{i=1}^{n-2k^2}\|\hat{\phi}_i-\phi_i\|^2}\right)\leq\sqrt{\left(\frac{\gamma}{9}\right)^2(n-2k^2)}<\left(\frac{\gamma}{9}\right)n
$$

with probability greater than $(1-\frac{\delta}{4})$, where $\hat{\theta}^{''}$ is the estimate of $\theta^{''}$.

Note that the dimension of $\theta^{''}$ is $k(n-2k^2)(2+2k^2)$, 
where each $P_{S_i}(\theta)\setminus P_S(\theta)$ encompasses $i$'th coordinate for each component mean, $i$'th diagonal entry for each component covariance matrix and $2k^2$  extra off diagonal entries for each component covariance matrix. These parameters represent the diagonally shaded region in Figure \ref{fig:reconstruction}.

\noindent \textbf{2b: Estimating the remaining entries of covariance matrices}

To estimate the the remaining parameters of the covariance matrices we need to estimate entries corresponding to pairs of coordinates $(i,j)$ when both $i$ and $j$ are not in $S$.
We take the projection onto $S_{ij} = span(S, e_i, e_j)$. It can be seen as before that radius of identifiability does not decrease going from $S$ to $S_{ij}$.
By applying Corollary~\ref{cor:general_case_1}, we will show that the $ij$'th entry of covariance matrices can be estimated.  Since there are $n-2k^2\choose 2$ such projections, we repeat this procedure $n-2k^2\choose 2$ times, each time 
we project $p_{\theta}$ onto appropriate $S_{ij}$ and invoke the algorithm of 
Corollary~\ref{cor:general_case_1}, (setting the precision parameter to $\frac{\gamma}{9}$) using a sample of size $\poly\left(\frac{1}{\gamma},\frac{1}{(\delta/4n^2)},B\right)$. Clearly this sample size is polynomial in $n$.
This ensures that, each time we get an estimate $\widehat{P_{S_{ij}}(\theta)}$ such that with probability at least $1-\frac{\delta}{4},~\|\widehat{P_{S_{ij}}(\theta)}-P_{S_{ij}}(\theta)\|\leq\frac{\gamma}{9}$.
Since 
$P_S(\theta)\subset P_{S_{ij}}(\theta)$ in each case and there are $n-2k^2\choose  2$ such cases, letting $\psi_t, t=1,\ldots,{n-2k^2\choose 2}$ to be the extra parameters in each case , we have for each $t$, $\|\hat{\psi}_t-\psi_t\|\leq\frac{\gamma}{9}$
with probability greater than $(1-\frac{\delta}{4})$, where $\hat{\psi}_t$ is the estimate of $\psi_t$. 
As before estimates of these extra parameters can be uniquely associated to the parameters of the component Gaussian distributions estimated in Step 2.
Letting $\theta^{'''}$ to be the $k{n-2k^2\choose 2}$ covariance parameters that have not been estimates in the previous steps, we have  $\theta^{'''}\subset\cup_{t=1}^{n-2k^2\choose 2}\psi_t$, and in particular, 
$$
\|\hat{\theta}^{'''}-\theta^{'''}\|\leq\left(\sqrt{\sum_{t=1}^{n-2k\choose 2}\|\hat{\psi}_t-\psi_t\|^2}\right)\leq\sqrt{\left(\frac{\gamma}{9}\right)^2{n-2k^2\choose 2}}<\left(\frac{\gamma}{9}\right)n
$$
where $\hat{\theta}^{'''}$ is the estimate of $\theta^{'''}$, with probability greater than $(1-\frac{\delta}{4})$. 
The parameters represented by $\theta^{'''}$ are shown in the vertically shaded region of Figure \ref{fig:reconstruction}. 

%

In Step 1 we need to invoke Theorem~\ref{thm:existence_of_poly_algo} $n\choose 2k^2$ times. In step 2 we need to invoke Corollary~\ref{cor:general_case_1} $1+(n-2k^2)+{n-2k^2\choose 2}$ times. Thus total invocation of Theorem~\ref{thm:existence_of_poly_algo} and Corollary~\ref{cor:general_case_1} combined is $\poly(n)$.
Now note that if $\epsilon<\fR(\theta)$ then $\gamma=\frac{\epsilon}{n}$. On the other hand if $\epsilon\geq\fR(\theta)$ then $\gamma=\frac{\fR(\theta)}{n}\leq\frac{\epsilon}{n}$.

Since $\theta^{'}\cup\theta^{''}\cup\theta^{'''}=\theta$,  
the corresponding estimate (with a little abuse of notation)  $\hat{\theta}=\hat{\theta}^{'}\cup\hat{\theta}^{''}\cup\hat{\theta}^{'''}$, with probability greater than $(1-\delta)$, is within $\epsilon$ of $\theta$ only up to a permutation using a sample of size $\poly\left(n,\max\left(\frac{1}{\epsilon},\frac{1}{\fR(\theta)}\right),\frac{1}{\delta}, B\right)$.
\ep

\noindent\textbf{Proof of Theorem \ref{thm:identifiability_detection} :}

Theorem \ref{thm_identifiability}, guarantees the existence of a $2k^2$-coordinate plane $S$ such that when $p_{\theta}$ is projected onto $S$, the corresponding mixture $\pi_S(p_{\theta})$, parameterized by $P_S(\theta)$, satisfies that $\fR(P_S(\theta))\geq\fR(\theta)\frac{1}{n}$. Since $S$ is not known in advance, projecting $p_{\theta}$ on to all $n\choose 2k^2$, $2k^2$-coordinate planes, each time   invoking the algorithm of Theorem \ref{thm:existence_of_poly_algo} with a sample of size $\poly\left(\frac{1}{(\epsilon/3n)},\frac{1}{(\delta/n^2)},B\right)$ and using union bound ensures that for each $2k^2$-coordinate plane $T$, Theorem \ref{thm:existence_of_poly_algo} produces a value of parameters $\widehat{P_T(\theta)}$ such that 
 $\widehat{P_T(\theta)}\in\mathcal{N}\left(P_T(\theta),\frac{\epsilon}{3n}\right)$ with probability greater than $(1-\delta)$. Now for each such $2k^2$-coordinate plane $T$, Lemma \ref{lem_radius_of_identifiability} guarantees that $|\fR{\widehat{P_T(\theta)}}-\fR{P_T(\theta)}|\leq\frac{\epsilon}{3n}$. 
Thus there must exist at least one $2k^2$-coordinate plane (say $T_*$) such that, 
$\fR(\widehat{P_{T_*}(\theta)})\geq\fR(\theta)\frac{1}{n}-\frac{\epsilon}{3n}$. Thus,
$$
(\fR(\theta)\geq\epsilon)\Rightarrow \left(\fR(\widehat{P_{T_*}(\theta)})\geq\frac{2\epsilon}{3n}\right)
$$
The desired algorithm now works as follows. For each of the $n\choose 2k^2$ values of parameters $\widehat{P_T(\theta)}$ outputted by Theorem~\ref{thm:existence_of_poly_algo},  we compute $\fR(\widehat{P_T(\theta)})$ using Equation \ref{eq:radius_identifiability}. Now set $\fR_*=\max_T\fR(\widehat{P_T(\theta)})$. 
If $\fR_*<\frac{2\epsilon}{3n}$ then output $\fR(\theta)<\epsilon$ otherwise output 
$\fR(\theta)\geq\epsilon$.
\ep

\noindent\textbf{Proof of Theorem~\ref{thm_identifiability} :}

%
Lemma \ref{lem:unique_mean} establishes the existence of a $k^2$-coordinate plane $S_1$, such that 
$\forall_{i,j}, ~\|P_{S_1}(\mu_i)-P_{S_1}(\mu_j)\|^2\geq\|\mu_i-\mu_j\|^2\frac{1}{n}>\|\mu_i-\mu_j\|^2\frac{1}{n^2}$. Similarly Lemma \ref{lem:unique_var} establishes 
the existence of a $k^2$-coordinate plane $S_2$, such that 
$\forall_{i,j}~\|P_{S_2}(\Sigma_i)-P_{S_2}(\Sigma_j)\|^2\geq\|\Sigma_i-\Sigma_j\|^2\frac{1}{n^2}$. Taking the span of these two planes produces a $2k^2$-coordinate plane $S=span(S_1, S_2)$,  such that
$$\min_{i\neq j}\left(\|P_S(\mu_i)-P_S(\mu_j)\|^2+\|P_S(\Sigma_i)-P_S(\Sigma_j)\|^2\right)\geq\min_{i\neq j}\left(\|\mu_i-\mu_j\|^2+\|\Sigma_i-\Sigma_j\|^2\right)\frac{1}{n^2}$$ 

Note that radius of identifiability of $\pi_S(p_{\theta})$, parameterized by $P_S(\theta)$, is given by,
\begin{eqnarray*}
(\fR(P_S(\theta)))^2  &=& \min\left(\frac{1}{4}\min_{i\neq j}\left(\|P_S(\mu_i)-P_S(\mu_j)\|^2+\|P_S(\Sigma_i)-P_S(\Sigma_j)\|^2\right),\min_i w_i^2\right)\\
&\geq& \min\left(\frac{1}{4n^2}\min_{i\neq j}\left(\|\mu_i-\mu_j\|^2+\|\Sigma_i-\Sigma_j\|^2\right),\min_i w_i^2\right)\\
\end{eqnarray*}
where the inequality follows from the fact that $\forall a_1, a_2, b, (a_1\leq a_2)\Rightarrow( \min(a_1,b)\leq\min(a_2,b))$.

\noindent\textbf{case 1:} $\frac{1}{4}\min_{i\neq j}\left(\|\mu_i-\mu_j\|^2+\|\Sigma_i-\Sigma_j\|^2\right)\leq\min_i w_i^2$\\

Here $(\fR(\theta))^2=\frac{1}{4}\min_{i\neq j}\left(\|\mu_i-\mu_j\|^2+\|\Sigma_i-\Sigma_j\|^2\right)$ and\\
 $(\fR(P_S(\theta)))^2\geq\left(\frac{1}{n^2}\right)\frac{1}{4}\min_{i\neq j}\left(\|\mu_i-\mu_j\|^2+\|\Sigma_i-\Sigma_j\|^2\right)=\left(\frac{1}{n^2}\right)(\fR(\theta))^2$.

\noindent\textbf{case 2:} $\frac{1}{4}\min_{i\neq j}\left(\|\mu_i-\mu_j\|^2+\|\Sigma_i-\Sigma_j\|^2\right)>\min_i w_i^2$\\

Here $(\fR(\theta))^2=\min_i w_i^2$.

If $\frac{1}{4}\min_{i\neq j}\left(\|\mu_i-\mu_j\|^2+\|\Sigma_i-\Sigma_j\|^2\right)>\min_i w_i^2>\frac{1}{4n^2}\min_{i\neq j}\left(\|\mu_i-\mu_j\|^2+\|\Sigma_i-\Sigma_j\|^2\right)$ then 
\begin{eqnarray*}
(\fR(P_S(\theta)))^2  &\geq& \min\left(\frac{1}{4n^2}\min_{i\neq j}\left(\|\mu_i-\mu_j\|^2+\|\Sigma_i-\Sigma_j\|^2\right),\min_i w_i^2\right)\\
&=&\frac{1}{n^2}\left(\frac{1}{4}\min_{i\neq j}\left(\|\mu_i-\mu_j\|^2+\|\Sigma_i-\Sigma_j\|^2\right)\right)\\
&>&\frac{1}{n^2}\min_i w_i^2=(\fR(\theta))^2\frac{1}{n^2}
\end{eqnarray*}

On the other hand if $\min_i w_i^2\leq\frac{1}{4n^2}\min_{i\neq j}\left(\|\mu_i-\mu_j\|^2+\|\Sigma_i-\Sigma_j\|^2\right)$ then,\\
$(\fR(P_S(\theta)))^2\geq\min_i w_i^2=(\fR(\theta))^2>\left(\frac{1}{n^2}\right)(\fR(\theta))^2$.
\ep


\section{Moment Concentration} 
\begin{lemma}\label{lem:moment_concentration}
Let $p_{\theta}, \theta\in\Theta\subset\mathbb{R}^m$ be a $m$-parametric family of probability
distributions in $\mathbb{R}^l$ where $\Theta$ is contained in a ball of radius $B$ in
$\mathbb{R}^m$ and let $X_1, X_2,\ldots,X_M$ be iid random vectors drawn from $p_{\theta}$.
Suppose the moments $M_{i_1\ldots i_l}(\theta)=\int x_1^{i_1}\ldots x_l^{i_l}dp_{\theta}$ and
the corresponding empirical moments $\hat{M}_{i_1\ldots i_l}(\theta)=\frac{\sum_{i=1}^M X_{i,1}^{i_1}\ldots X_{i,l}^{i_l}}{M}$
are lexicographically ordered as $M_1(\theta), M_2(\theta),\ldots$ and
$\hat{M}_1(\theta), \hat{M}_2(\theta),\ldots$ respectively. Then given any positive integer $N$, and sample size
$M>\frac{C NB^{2\lceil\frac{N}{l}\rceil}}{\epsilon^2\delta}$ where $C$ is a constant, 
for any $\epsilon>0$ and $0<\delta<1,~~|\hat{M}_i(\theta)-M_i(\theta)|\leq\epsilon$  for all $i\leq N$ with probability greater than $1-\delta$.
\end{lemma}
\bp
For any $i\leq N$, let $M_i(\theta)=\int x_1^{a_1(i)}x_2^{a_2(i)}\ldots x_l^{a_l(i)}dp_{\theta}$, where $a_j(i)$ is a function of $i$ for $j=1,2,...,l$. Let $f_i:\mathbb{R}^l\rightarrow \mathbb{R}$ be a function defined as $f_i(x)=x_1^{a_1(i)}x_2^{a_2(i)}\ldots x_l^{a_l(i)}$. 
For any random vector $X$ distributed according to $p_{\theta}$, we have $\mathbb{E}[f_i(X)]=M_i(\theta)$. The empirical counterpart is defined 
as $\frac{\sum_{j=1}^M f_i(X_j)}{M}=\hat{M}_i(\theta)$. Note that $\mathbb{E}\left(\frac{\sum_{j=1}^M f_i(X_j)}{M}\right)=\mathbb{E}[f_i(X)]$.
Now

$\var\left(\frac{\sum_{j=1}^M f_i(X_j)}{M}\right)=\frac{\var\left(f_i(X)\right)}{M}=\frac{1}{M}\mathbb{E}\left(f_i(X)-\mathbb{E}(f_i(X))\right)^2$\\
$~~~~~~~~~~~~~~~~~~~~~~~=\frac{1}{M}\left(\mathbb{E}\left([f_i(X)]^2\right)-\left(\mathbb{E}[f_i(X)]\right)^2\right)$\\
$~~~~~~~~~~~~~~~~~~~~~~~\leq\frac{1}{M}\left\{\mathbb{E}\left([f_i(X)]^2\right)\right\}$\\
$~~~~~~~~~~~~~~~~~~~~~~~=\frac{1}{M}\int x_1^{2a_1(i)}x_2^{2a_2(i)}\ldots x_l^{2a_l(i)}dp_{\theta}$\\
$~~~~~~~~~~~~~~~~~~~~~~~\leq \frac{C B^{2\lceil\frac{N}{l}\rceil}}{M}$\\
where the last inequality follows from the fact that when the moments are lexicographically ordered, 
for any $i\leq N$, the maximum degree 
of the polynomial $x_1^{a_1(i)}x_2^{a_2(i)}\ldots x_l^{a_l(i)}$ is at most $\lceil\frac{N}{l}\rceil$.

Now applying Chebyshev's inequality we get,\\
$P\left(|\hat{M_i}(\theta)-M_i(\theta)|>\epsilon\right)=P\left(\left|\frac{\sum_{j=1}^M f_i(X_j)}{M}-\mathbb{E}[f_i(X)]\right|>\epsilon\right)$\\
$~~~~~~~~~~~~~~~~~~~~~~~~~~~~~~~~~\leq\frac{\var\left(\frac{\sum_{j=1}^M f_i(X_j)}{M}\right)}{\epsilon^2}\leq\frac{C B^{2\lceil\frac{N}{l}\rceil}}{M\epsilon^2}$

Upper bounding the last quantity by $\frac{\delta}{N}$ and using union bound yields the desired result.
\ep

\end{document}